%% file: main.tex
\documentclass[preprint,12pt,authoryear]{elsarticle}

\usepackage{amssymb}
\usepackage{lipsum}
\usepackage{url}
\usepackage{amsmath} 
\usepackage{color}
\usepackage{float}
\usepackage{multirow} 
\usepackage{algorithm}
\usepackage{algorithmic}
\usepackage{tikz}
\usepackage{subcaption}
\usepackage{booktabs}
\usetikzlibrary{calc}

\usepackage[utf8]{inputenc}
\usepackage{graphicx}
\usepackage{booktabs}
\usepackage{colortbl}
\usepackage{float}
\usepackage{lineno}
\usepackage[dvipsnames]{xcolor}
\usepackage{stackengine}

\journal{Elsevier}

\begin{document}

\begin{frontmatter}



\title{A geometric and deep learning reproducible pipeline for monitoring floating anthropogenic debris in urban rivers using in situ cameras}


\author[LIVE]{Gauthier Grimmer}
\author[LIVE]{Romain Wenger\corref{cor1}}
\author[CAMUS,TUINH04]{Clément Flint} 
\author[IRIMAS,DSAI]{Germain Forestier} 
\author[LIVE]{Gilles Rixhon} 
\author[LIVE]{Valentin Chardon} 
\affiliation[LIVE]{organization={LIVE UMR 7362 CNRS, University of Strasbourg},
            addressline={3 rue de l'Argonne}, 
            city={Strasbourg},
            postcode={67000}, 
            country={France}}

\affiliation[CAMUS]{organization={Centre Inria de l'Université de Lorraine},
            addressline={615 rue du Jardin Botanique}, 
            city={Villers-lès-Nancy},
            postcode={54600}, 
            country={France}}

\affiliation[TUINH04]{organization={Technical University of Munich, Campus Heilbronn},
            addressline={Bildungscampus 2},
            city={Heilbronn},
            postcode={74076},
            country={Germany}}

\affiliation[IRIMAS]{organization={IRIMAS UR 7499, University of Haute-Alsace},
            addressline={12 rue des Frères Lumière}, 
            city={Mulhouse},
            postcode={68000}, 
            country={France}}

\affiliation[DSAI]{organization={Data Science and Artificial Intelligence (DSAI), Monash University},
            city={Melbourne},
            country={Australia}}
            
\cortext[cor1]{Corresponding author. Email address: romain.wenger@live-cnrs.unistra.fr (R.Wenger)}

\begin{abstract}
The proliferation of floating anthropogenic debris in rivers has emerged as a pressing environmental concern, exerting a detrimental influence on biodiversity, water quality, and human activities such as navigation and recreation. 
The present study proposes a novel methodological framework for monitoring the aforementioned waste, utilising fixed, in-situ cameras. 
The study innovatively combines (i) a reproducible experimental framework for investigating dataset biases (i.e., negative images, temporal leakage, class weighting), with (ii) an evaluation of object detection under bank-mounted, oblique-view conditions, and (iii) an interpretable monocular geometric pipeline for estimating object dimensions from detections.
These models are tested in a range of environmental conditions and learning configurations, including experiments on biases related to data leakage. 
Furthermore, a geometric model is implemented to estimate the actual size of detected objects from a 2D image. 
This model takes advantage of both intrinsic and extrinsic characteristics of the camera. 
The findings of this study underscore the significance of the dataset constitution protocol, particularly with respect to the integration of negative images and the consideration of temporal leakage. 
In conclusion, the feasibility of object dimension estimation using projective geometry coupled with regression corrections is demonstrated. 
This approach paves the way for the development of robust, low-cost, automated monitoring systems for urban aquatic environments.
\end{abstract}




\end{frontmatter}




\section{Introduction}
\label{sec:introduction}

Several million tonnes (1.15–2.41 Mt) of anthropogenic debris are transported annually to the oceans via hydrosystems \citep{lebreton2017river, gonzalez2021floating, palmas2022rivers}.
Their presence in rivers results from both anthropogenic activities—such as poor waste management, illegal dumping, large spatial disparity in waster-water-treatment or sprawling urbanisation \citep{van2020plastic, schoneich2020wasting, de2021quantifying}—and natural processes like aerial deflation and surface runoff induced by rainfall \citep{van2020plastic}.
This accumulation can be particularly significant downstream of urban areas, due to high population density and socio-economic activities significantly contribute to various debris inputs into hydrosystems \citep{rech2014rivers, van2020plastic, gomez2022learning}.
Anthropogenic debris—including plastics, metals, and glass \citep{hanke2013guidance, suteja2025spatial}—represent a major environmental concern, especially in the context of urban river management.
Plastics alone accounts for approximately 80\% of these debris \citep{rech2014rivers, gonzalez2021floating}, making it a pervasive pollutant affecting all ecosystems.

In rivers, macroplastics ($>$ 5 mm; \citealp{gonzalez2021floating}) degrade into microplastics ($<$ 5 mm; \citealp{skalska2020riverine}) through biochemical (i.e. photodegradation, oxidation, biodegradation) and mechanical (i.e. grinding and repeated impacts during transport with water and sediment) fragmentation \citep{liro2023macroplastic}.
Aligned with the One Health approach \citep{prata2021one, bois2025rethinking}, macroplastics harm wildlife (i.e. entanglement, intestinal blockage; \citealp{waring2018plastic, alabi2019public}) and promote bank overflow and flooding \citep{van2020plastic} eventually resulting in river morphology alterations.
The average concentration of 6.7 particles of microplastics per cubic metre of water in European rivers \citep{ghiglione2023mission, landebrit2024small} represents a growing environmental and public health concern.
Furthermore, microplastics interfere with natural ecosystem functions (i.e. nutrient flow, species reproduction, etc.; \citealp{prata2021one}) and are easily found in the trophic chain through the consumption of fish \citep{neves2015ingestion} or other products containing plastics \citep{kumar2022micro}.
Plastics directly impact health due to additives like Bisphenol A and Persistent Organic Pollutants (POPs), associated with neurological and reproductive issues \citep{alabi2019public}.
This underlines the need to monitor riverine debris to reduce macro- and microplastics input in hydrosystems.

Riverine anthropogenic debris and their spatial dynamics have been studied so far through direct monitoring methods.
The latter include : (i) Global Positioning System (GPS) \citep{tramoy2019assessment} or Radio Frequency Identification (RFID) trackers \citep{chardon2025simple}, (ii) debris quantification via visual observation \citep{castro2019macro}, (iii) net sampling from bridges or boats \citep{hurley2023measuring}, or (iv) passive sampling (i.e. use of existing infrastructure like hydraulic structures and floating debris traps; \citealp{van2020plastic}).
Although these methods can be efficient at the local scale, they fail, at the catchment scale, to continuously monitor debris flux \citep{hurley2023measuring}.
These inherent limitations call for new automated approaches enabling continuous monitoring and the production of reproducible results.

In this respect, camera-acquired digital images have been recently used to tackle issues related to anthropogenic debris \citep{van2020automated, donal2023automated, zhang2023yolov5} because they allow continuous recording and thereby capturing data variability (i.e. debris diversity, water turbidity, weather conditions; \citealp{jia2023deep, tata2021robotic}).
In connection with riverine debris monitoring, deep learning (DL) methods have been increasingly used.
\cite{van2020automated} and \cite{kataoka2024instance} took advantage of cameras mounted on bridges characterised by vertical views.
\cite{maharjan2022detection} used cameras onboard UAVs.
Images captured by these methods are then processed with convolutional neural networks (CNNs)—such as Faster R-CNN \citep{van2020automated}, YOLOv5-s (You Only Look Once) \citep{armitage2022detection} or YOLOv8 \citep{kataoka2024instance}—achieving high detection performance on riverine debris.

{Multi-scale detection frameworks have been proposed to capture objects of varying sizes in high-resolution imagery. 
Additionally, multi-branch architectures improve feature representation through complementary processing pathways.
These approaches have shown strong performance in remote sensing contexts (i.e. satellite imagery —\citealp{khan2022unified, khan2023multi}).

Three main limitations remain concerning camera-based debris studies.

Firstly, the majority of works focused on the performance of detection, without explicitly addressing the influence of dataset construction and evaluation protocols. 
In particular, the role of negative images and the presence of temporal leakage between training and test data are rarely investigated, despite the fact that they may significantly affect model generalization.

Secondly, a significant number of studies are constrained by their monitoring scale, depending either on bridge-mounted cameras fixed to a specific spot or UAV-based acquisitions, which are subject to inherent temporal limitations.
Consequently, there is a lack of bank-mounted camera installations, which are more readily deployable and do not depend on existing anthropogenic infrastructure. 
Although these configurations generate pronounced perspective effects and more intricate observation conditions, they align more closely with the operational requirements peculiar to long-term monitoring systems.

Thirdly, the majority of existing approaches are limited to detection ouptputs, with little integration of physically interpretable measurements such as object dimensions derived from monocular imagery.

The present study proposes three main contributions to overcome these limitations
It first relies on introduces a reproducible experimental framework to assess dataset-related biases, particularly the effects of negative images and temporal leakage on model performance. 
It also evaluates debris detection under bank-mounted, oblique-view conditions; this remain less explored but is more representative of real-world deployments. 
Finally, it proposes an interpretable monocular geometric pipeline to estimate object dimensions from RGB imagery using camera parameters and regression-based correction.

\section{Methodology}
\subsection{Study area}

The natural reserve of the \textit{Robertsau} and \textit{Wantzenau} forest massif is located in the northen part of the Greater Strasbourg in the Rhine floodplain (E. France, Fig.\ref{fig2:study_area}).
A protected natural area was established in 2020, and covers a total area of 710 ha. 
A wide breadth of natural environments, including deciduous forests, waterways, arable lands, and wetlands makes it peculiar.
The forest massif of the reserve is considered representative of lowland alluvial environments in temperate regions \citep{schnitzler1994conservation}. 
Although the neighbouring Rhine River has now been dammed up (Fig.\ref{fig2:study_area}), it still exerts a significant influence, notably by maintaining the hydrological dynamics of the riparian zone. 
This diversity of habitats hosts a wide variety of flora and fauna, including species such as the European beaver, the European eel, the elm or the superb carnation \citep{RNF_Robertsau_Wantzenau}.

The Steingiessen is a 4.5-kilometre-long sinuous watercourse and a functional defluent of the Ill River, situated within the Greater Strasbourg hydrographic system (Fig.\ref{fig2:study_area}). 
Crucially, its location immediately downstream of a major urban centre makes it a representative river for studying the transfer of anthropogenic pressures to peri-urban environments. 
While the Steingiessen drains a protected nature reserve that demands stringent environmental preservation, its hydrological connection to the Ill River renders it vulnerable to urban-sourced pollution and debris. 
This dual character—as both a high-value ecological corridor and a receptacle for upstream discharge—contextualises the site’s strategic importance.
Downstream, the system joins the Rhine drainage canal and eventually the Rhine River itself, bypasses the Gambsheim hydroelectric dam via the Ill diversion canal.

\begin{figure}[H]
    \centering
    \includegraphics[width=1\linewidth]{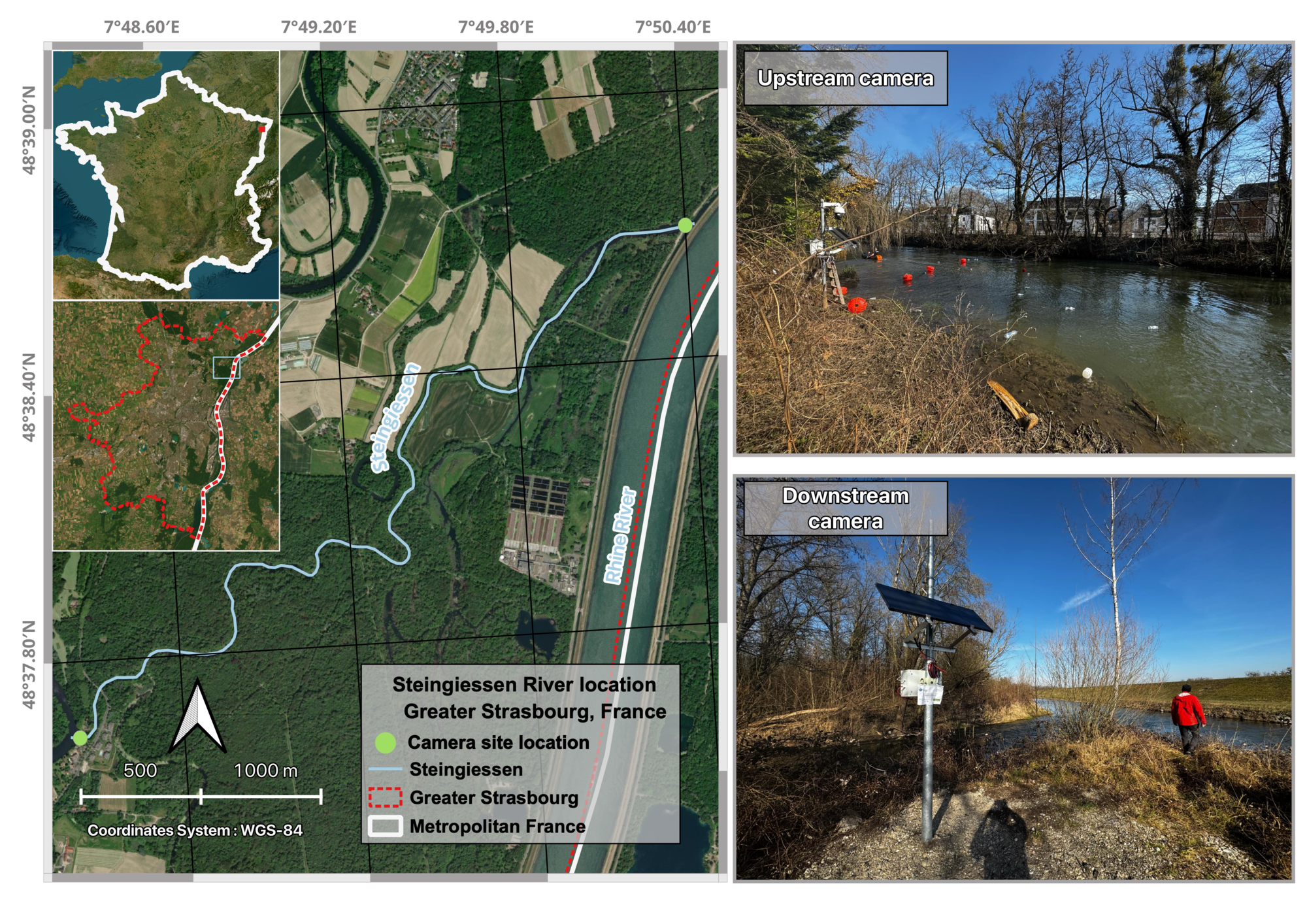}
    \caption{Location of Steingiessen River and inlet and outlet cameras (north of Greater Strasbourg)}
    \label{fig2:study_area}
\end{figure}

\subsection{Data}
\subsubsection{In situ data collection}

For this study, two pan-tilt-zoom (PTZ) cameras were used to capture model training data. 
These cameras were positioned in the upstream- and downstream-most reach of the Steingiessen River (Fig.\ref{fig2:study_area}).

To build a controlled detection dataset under field conditions, anthropogenic debris and wood were manually introduced into the river during dedicated acquisition sessions. 
On March 4, 2025, between 10 a.m. and 12 p.m., several types of anthropogenic debris (glass bottles, plastic bags, polystyrene, plastic bottles) as well as natural debris (wood) were released upstream of the first camera. 
At the same time, kayak passages were performed to create examples of non-debris floating objects. 
The same protocol was repeated from 3 p.m. to 4 p.m. for the downstream camera.

To avoid any additional environmental contamination, all manually introduced objects were recovered during the experiment by a team member navigating downstream by kayak. 
This acquisition protocol made it possible to construct a controlled benchmark while remaining in real river conditions.

In addition, although the positive acquisitions were conducted during a single experimental day, the image collection strategy more broadly aimed to capture the study site under different illumination conditions, in order to better reflect the visual variability encountered in practice. 
However, rainy conditions could not be included in the acquisition protocol.

The final dataset contains a total of 6,013 images.

\subsubsection{Image labelling and ground truth data creation}

The first step in object detection is data labelling (Fig.\ref{fig1:employed_methodology}). 
The images are manually annotated using the \textit{LabelImg} tool \citep{tzutalin_labelimg_2015}, with the annotation process consisting of the creation of bounding boxes around objects (Fig.\ref{fig3_annoted_data}) \citep{aarnink2025automatic}.
The annotation process was made according to following typology : (1) anthropogenic debris, (2) natural debris, and (3) non-debris material (floating boats).
The initial dataset contains : 10,873 anthropogenic debris, 754 natural debris, and 1,010 non-debris materials.

\begin{figure}[H]
    \centering
    \includegraphics[width=1\linewidth]{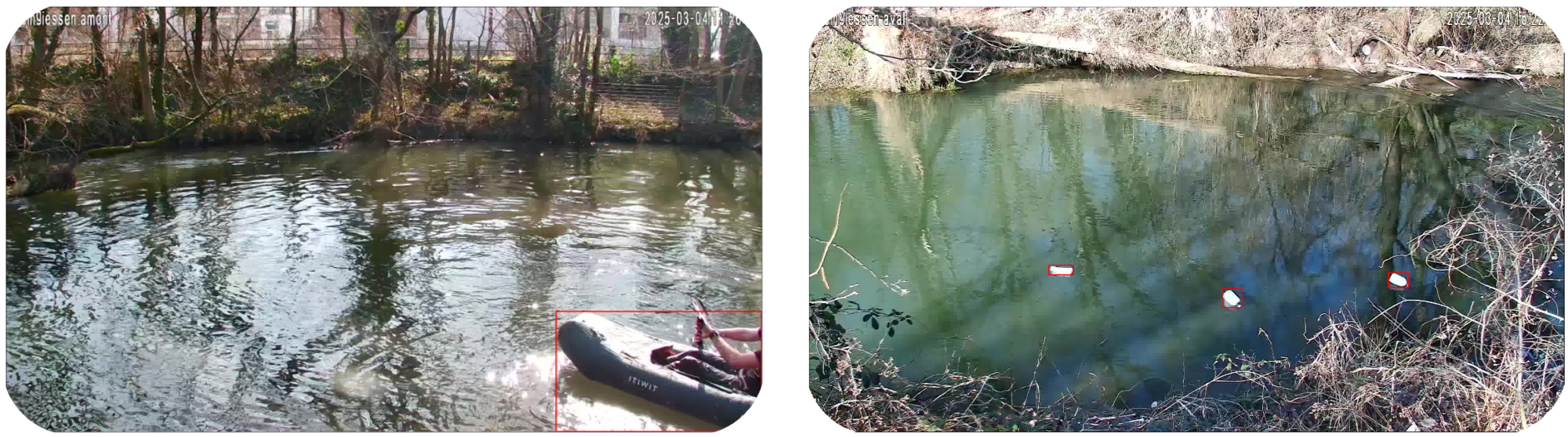}
    \caption{Data acquired on March 4, 2025 upstream (left) and downstream (right) of the Steingiessen River. Bounding boxes on anthropogenic debris and non-debris materials. Bounding boxes have been manually digitized.}
    \label{fig3_annoted_data}
\end{figure}

\subsubsection{Negative images acquisition}

The incorporation of negative images, corresponding to images without objects, aims to test the hypothesis that such data reduces model bias \citep{malagon2009object}.
Indeed, the presence of objects on the river surface can be overestimated due to other variable features inherent to real-life conditions (i.e., luminosity, reflections, eddies).

Negative images were selected for the month of February 2025. 
Variables relating to sunshine were used to select days with either sunny or cloudy weather conditions. They are: 
(1) the daily insolution duration (INST);
(2) daily global irradiance (GLOT);
(3) and fraction of sunshine in relation to day length (SIGMA). 
The variables were normalised to determine the days of interest (Fig.\ref{fig4:images_neg}).
Consequently, 10 February 2025 was identified as a day characterised by low levels of sunshine, while 20 February 2025 was considered a day with clear skies.
In addition, negative images were acquired from the upstream and downstream cameras. 

A total of 14,030 negative images were acquired: 7,098 and 6,932 images were recorded on 10 and 20 February, respectively (Fig.\ref{fig4:images_neg}).

\begin{figure}[H]
    \centering
    \includegraphics[width=1\linewidth]{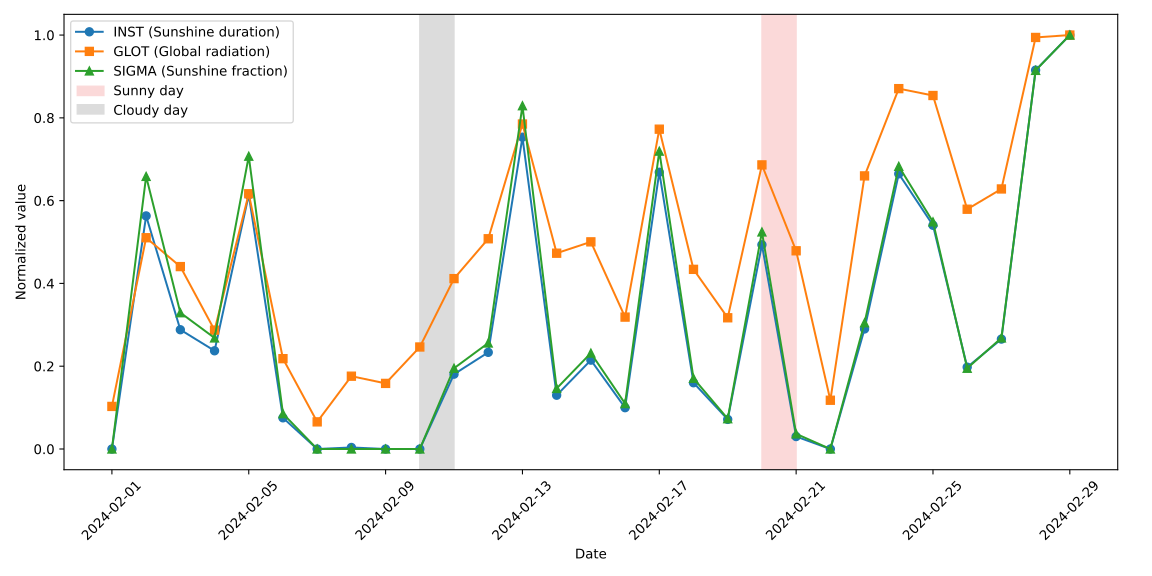}
    \caption{Evolution of Daily Insolation Duration (INST), Daily Global Radiation (GLOT) and fraction of sunshine in relation to day length (SIGMA) during February 2025 at the Strasbourg - Entzheim weather station (France) (“\textit{Données climatologiques de base - quotidiennes}”, Météo-France, 2025). Grey and pink strips refer to cloudy and sunny weather conditions, respectively.}
    \label{fig4:images_neg}
\end{figure}

\subsection{Data leakage}
\label{section2.3}

Splitting randomly image data into three subsets (train,  validation, test) can lead to data leakage. 
This occurs when a subset of the training data is also used in the test data \citep{babu2024improving}. 
The data leakage problem can distort the actual performance of the object detection model by over- or underestimating performance measures \citep{babu2024improving}. 
This error can occur in various ways, notably in features and temporal data \citep{john2025problematic}.
In our case, the use of data in unvaried forms (i.e. the majority of anthropogenic debris are plastic bottles) and temporal data can lead to data leakage.
To limit this bias, we assemble image data into groups that contain images with the same temporal scene and, therefore, the same debris.

\subsubsection{Extraction of image features}

Image data contains substantial redundancy, so reducing its dimensionality through feature extraction is necessary to group it effectively \citep{figueiredo2024analyzing}.  
As a first step, a pre-trained YOLOv8 (You Only Look Once) model was used to extract visual features from images in its $21^\text{st}$ back layer (Fig.\ref{fig5:yolo_archi}).
This generates a 256-length embedded vector. 
This means that the $21^\text{st}$ layer of the YOLOv8 encoder has identified 256 visual features in each image. 
In addition, we extracted features from the labelling annotation for each image, such as the number of objects per class and information about the bounding box of the annotated objects.
The timestamps were extracted and added to the final vector, creating an embedded vector for 6,013 images, 261 in length.
Given the size of the data, enhanced analysis requires reducing it to two dimensions. 

\subsubsection{Dimension reduction using t-Distributed Stochastic Neighbor Embedding}

t-SNE (t-Distributed Stochastic Neighbor Embedding) is a non-linear, unsupervised method to map data into reduced dimensions (generally 2 dimensions) \citep{anowar2021conceptual} while preserving the significant structure of the original data \citep{figueiredo2024analyzing}.
It reduces dimensionality by converting  the Euclidian distance between high-dimensional data points into conditional probabilities representing similarities \citep{oliveira2018use}, calculated as:

\begin{align}
p_{a|b} = \frac{\exp\left(-\frac{\|x_b - x_a\|^2}{2\sigma^2}\right)}{\sum_{a \ne k} \exp\left(-\frac{\|x_k - x_a\|^2}{2\sigma^2}\right)}
\label{eq:t-SNE_proba}
\end{align}

where $x_a$ and $x_b$ are two data points.
It calculates how close is $x_a$ is from $x_b$ considering a Gaussian distribution around $x_b$ with a given variance $\sigma^2$\citep{anowar2021conceptual}.
When the probabilities of distributions are calculated, the aim is to minimize the difference between two points. 
The function to be minimized is the sum of the Kullback-Leibler divergences at all points with a gradient descent \citep{oliveira2018use}:

\begin{align}
E = \sum_{a} KL(P_a \parallel Q_a) = \sum_{a} \sum_{b} p_{b|a} \log \frac{p_{b|a}}{q_{b|a}},
\label{eq:t-SNE_minim}
\end{align}

where $P_a$ represents the conditional probability distribution measuring the similarity between the point $x_a$ and all other data points in the original space, while $Q_a$ is the corresponding conditional probability distribution computed from the projected point $y_a$ in the low-dimensional space \citep{oliveira2018use}. 
Therefore, t-SNE was calculated with a perplexity of 30 and a learning rate of 200.

\subsubsection{DBSCAN clustering}

Clustering partitions data into groups based on meaningful similarity measures, aiming to achieve high intra-group homogeneity and distinct inter-group differences \citep{wegmann2021review}.
After dimensionality reduction, clustering is used to group images from the same scene.
The Density-based Spatial Clustering of Application with Noise (DBSCAN) algorithm was chosen to cluster images \citep{figueiredo2024analyzing}.
This algorithm detects clusters as high-density area that can spread out in any direction, allowing the identification of arbitrary-shaped groups, outliers, and noise. \citep{wegmann2021review}.
DBSCAN was applied with an eps of 5 and a minimum sample of 10.

 \subsubsection{Clustering validation}

The clustering performance was assessed using the DBCV (Density-Based Clustering Validation) index.
This metric combines local density with structural separation to evaluate cluster quality. 
Initially, the core distance for each point is calculated, representing the inverse of its local density. 
Subsequently, a minimum spanning tree is employed to capture the internal structure for each cluster. 
The subsequent extraction of density sparseness and density separation is achieved through this process. 
These components are used to compute a metric that evaluates how dense and well-separated the clusters are \citep{wegmann2021review}. 
DBCV is calculated as : 

\begin{align}
DBCV(C) = \sum_{i=1}^{l} \frac{|C_i|}{|O|} \cdot V_C(C_i)
\label{eq:DBCV}
\end{align}

where \( C \) is the set of all clusters, \( l \) is the total number of clusters, \( |C_i| \) is the number of points in cluster \( C_i \), \( |O| \) is the total number of data points, and \( V_C(C_i) \) is the local validity score of cluster \( C_i \), based on its internal density and its separation from other clusters \citep{moulavi2014density}. 

\subsection{Dataset creation}

Three datasets were created. 
The first one was created by a random split of 80\% for the training set, 10\% for the validation set, and 10\% for the testing set.
It contains 6,013 images for a total of 12,637 objects of all classes. 
The training set contains 8,635 anthropogenic debris, 601 natural debris, and 824 non-debris material. 
The validation set contains 1119 anthropogenic debris, 76 natural debris,and 105 non-debris material.
Finally, the testing set contains 1119 anthropogenic debris, 77 natural debris and 81 non-debris material.

The second dataset was created using a random split (80/10/10) on clusters created by the data leak-free method (Section \ref{section2.3}). 
Each cluster is randomly placed in one of the three sets (training, validation and testing) to avoid separation.
Thus, the training set contains 8,978 anthropogenic debris, 424 natural debris and 465 non-debris material.
The validation set contains 475 anthropogenic debris, 145 natural debris, and 330 non-debris material. 
Finally, the testing set contains, 1,198 anthropogenic debris, 185 natural debris, and 215 non-debris material.

The third datasets was created to evaluate the impact of the proportion of negative images on model performance. 
Starting from the leak-free dataset (the second one), negative images were added to the training data at five different proportions: 0\%, 10\%, 20\%, 30\% and 40\%.
Consequently, the training set contains 0, 525, 1.174, 2.009 and 3.122 negative images, respectively. 
The validation sets contain 0, 67, 150, 257 and 399 negative images, respectively.
The testing set contains 0, 66, 148, 253 and 393 negative images, respectively.

\subsection{Deep learning approach}
\subsubsection{Selecting Convolutional Neural Network}

Convolutional neural networks (CNNs) are algorithms dedicated to processing data in the form of multiple matrices, such as signals and sequences, images, or videos. 
CNNs are composed of a cascading sequence of convolution and pooling layers, which extract essential features from the input data to perform class prediction \citep{lecun2015deep}. 
The convolution layer extracts the main features of data such as textures or edges by applying a convolutional kernel on it. 
In addition, the pooling layer performs a sampling action on the feature maps derived from the convolution layers to reduce their spatial dimensionality \citep{schwindt2024transfer} (Fig.\ref{fig5:yolo_archi}).

In the context of object detection methodologies, several CNN architectures are used, such as the R-CNN \citep{van2020automated} or YOLO algorithms \citep{aarnink2025automatic}.
R-CNN algorithms are based on a region-of-interest method.
First, they generate potential bounding boxes and classify them. 
The algorithm then performs a long-time post-processing \citep{aarnink2025automatic}, including removing detections and correcting bounding boxes by analysing other detections in the image \citep{girshick2014rich}.
In contrast, the YOLO model processes the entire image during the detection phase \citep{redmon2016you}. 
Because of this, the processing time of the YOLO algorithm is shorter, while its performance appears higher \citep{aarnink2025automatic, redmon2016you}. Consequently, the YOLO algorithm was selected for this study.

\begin{figure}[H]
    \centering
    \includegraphics[width=1\linewidth]{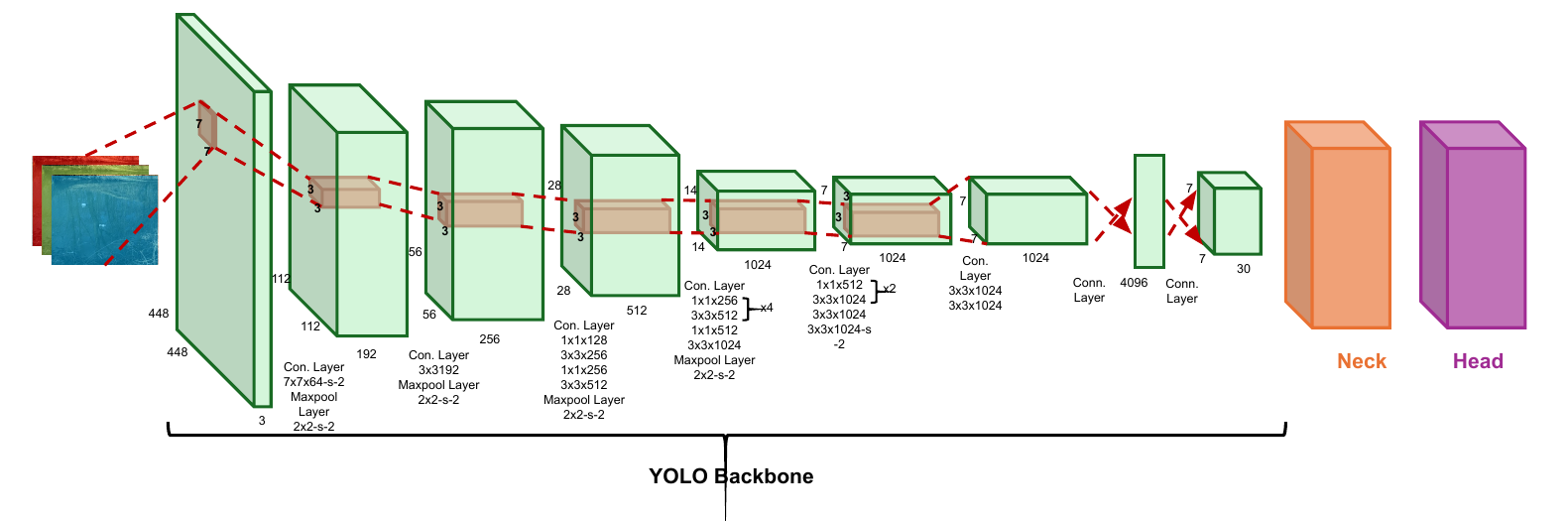}
    \caption{General architecture of a YOLO object detection model. The structure is divided into three main components: the \textbf{backbone}, responsible for extracting hierarchical features from the input image using a series of convolutional and pooling layers; the \textbf{neck}, which enhances feature aggregation across different scales (often using modules such as Path Aggregation Network (PANet) or Feature Pyramid Network (FPN) in recent versions); and the \textbf{head}, which performs final object detection by predicting bounding boxes, objectness scores, and class probabilities. While the figure reflects a simplified backbone resembling early YOLO versions, the general structure remains consistent across modern versions such as YOLOv5, YOLOv8, and YOLOv11, with architectural refinements aimed at improving speed and accuracy.}
    \label{fig5:yolo_archi}
\end{figure}

\subsubsection{Implementation details}

Data augmentation was performed \citep{de2021quantifying}.
The data augmentation parameters are as follows; vertical tilting on 50\% of images, horizontal tilting on 50\% of images, activation of mosaic augmentation, hue variation (0.015), saturation variation (0.7), brightness variation (0.4), perspective distortion (0.0005), random resizing on 50\% of images and random translation.
It (i) increases the amount of input data to the model \citep{kylili2019identifying}, (ii) adds variability to the dataset and increases its robustness \citep{aarnink2025automatic}.

The number of epochs is set to 200. 
All calculations were performed with an Nvidia GeForce GTX 1080ti graphics processing unit (GPU) with 11GB of VRAM. 
The training scenarios (see section \ref{subsect:yolo_train}) include YOLO nano and medium models.
Being limited in computing power, the nano models were run with a batch size of 32, while the medium variants of the models were run with a batch size of 8.
In addition, the image size is fixed at 1280 pixels.
This patch size enables the model to improve the detection of smaller objects and avoid misclassification due to lighting differences.
To optimize training, the learning rate (LR) was set at 0.001. 
To limit LR descent, a final learning rate (FLR) is set at 0.2, in conjunction with cosine annealing \citep{ultralytics2019training}.
A LR planner adjusts the LR according to a cosine curve, thus limiting the risk of exceeding the optimal LR value \citep{schwindt2024transfer}. 
Finally, a momentum (0.937) is introduced to facilitate the incorporation of previous gradients into new epochs \citep{ultralytics2019training}.

One of the recurrent challenges encountered during the training of DL algorithms pertains to the issue of overfitting.
This occurs when the model performs very well on the training dataset but has great difficulty generalizing to the test dataset \citep{ying2019overview}.
To limit this effect, an early-stopping callback, with a patience of 10 epochs, has been integrated into the hyperparameters of the trained models.
This approach serves to mitigate the impact of noise learning on the model's accuracy. 
So, if model accuracy falls continuously over 10 epochs, early-stopping stops learning and retains the best epoch of training.
In addition, an L2 (0.0005) regularization has been applied, notably to limit the weights of features with little significance in model learning \citep{ying2019overview}.

\subsubsection{Class imbalance mitigation}

Class imbalance is a well-known issue in CNN training, particularly when one class largely prevails over the others \citep{ghosh2024class}. 
This imbalance leads the model to prioritise the detection of the dominant class during training, resulting in reduced performance for minority classes and limited generalisation capacity \citep{pulgar2017impact, johnson2019survey}.

In standard object detection frameworks, the classification loss is typically based on the binary cross-entropy (BCE) loss, defined as:

\begin{align}
\mathcal{L}_{\text{BCE}} = - \sum_{c=1}^{C} \left[ y_c \log(\sigma(x_c)) + (1 - y_c)\log(1 - \sigma(x_c)) \right]
\end{align}

where $y_c$ is the ground truth label for class $c$, $x_c$ is the predicted logit, and $\sigma(\cdot)$ denotes the sigmoid activation function.

However, this formulation treats all classes equally, which can lead to suboptimal learning in imbalanced settings. 
To address this limitation, a weighted version of the BCE loss was introduced, incorporating class-specific weights:

\begin{align}
\mathcal{L}_{\text{WBCE}} = - \sum_{c=1}^{C} w_c \left[ y_c \log(\sigma(x_c)) + (1 - y_c)\log(1 - \sigma(x_c)) \right]
\end{align}

where $w_c$ denotes the weight associated with class $c$. This formulation increases the penalisation of misclassified samples from underrepresented classes, thereby enhancing their contribution during optimisation \citep{masko2015impact, gong2023survey}

Class weights were computed using an inverse frequency scheme based on the distribution of instances in the training dataset. This approach assigns higher weights to minority classes, ensuring that rare classes contribute more significantly to the loss function.

This experiment was conducted using the negative image configuration (from the 0\%, 10\%, 20\%, 30\%, and 40\% range) that yielded the highest performance metrics. 
The objective was therefore to assess whether class weighting provides additional benefits when applied on top of an optimised dataset design, rather than to isolate its effect independently.

\subsubsection{Experimental design for YOLO-based object detection}
\label{subsect:yolo_train}

Training was performed using transfer learning \citep{donal2023automated}. 
This method involves using artificial neural network models pre-trained on other datasets.
The architecture, weights, and biases of the models are then used to train the new model \citep{schwindt2024transfer}. 
Three variants of the YOLO algorithms were tested: YOLOv5, YOLOv8 and YOLOv11. 
These algorithms, dedicated to object detection, are models pre-trained on the COCO dataset \citep{lin2014microsoft} proposed by Microsoft, which contains 80 object classes. 
To compare the behaviour of several YOLO architectures under the experimental conditions of this study, ten training scenarios were tested.
The objective of this comparison is to identify relevant performance-efficiency trade-offs within the experimental framework under investigation, rather than to establish a universal ranking of object detection models for riverine debris monitoring.

\begin{itemize}

\item[] -- Baseline training on randomly split dataset: 
Six YOLO configurations (YOLOv5-n, YOLOv5-m, YOLOv8-n, YOLOv8-m, YOLOv11-n, YOLOv11-m) were trained on the initial dataset using a standard random split. All models were trained with the hyperparameters previously set.

\item[] -- Training on leak-free dataset:
To assess the impact of temporal leakage, the two selected architectures (YOLOv8-n and YOLOv11-m) were retrained on a second dataset created by random split on clusters from t-SNE and DBSCAN pipeline (Fig.\ref{fig1:employed_methodology}). It is trained with hyperparameters previously set.

\item[] -- Negative-image ablation experiments:
An ablation study was conducted on the leak-free dataset using YOLOv8-n and YOLOv11-m to evaluate the effect of background-only images. For each architecture, five models were trained using datasets containing 0\%, 10\%, 20\%, 30\%, and 40\% negative images. The objective was to analyze how different proportions of negative images influence detection performance.

\item[] -- Class imbalance mitigation via class weighting:
An additional experiment was conducted to evaluate the effect of class imbalance mitigation using class weighting. This experiment was performed on the leak-free dataset including the optimal proportion of negative images, identified in the previous ablation study. Class weights were incorporated into the classification loss to increase the contribution of underrepresented classes during training. The objective was to assess whether class weighting provides additional performance improvements when applied on top of an optimized dataset design.

\end{itemize}

\subsubsection{Model evaluation}

Indices such as precision, recall, mAP@50, and mAP@50-95 are used to assess the performances of the different models. 
To obtain these indices, the models are applied to the test sets. Recall (Eq. \ref{eq1:recall}) represents the proportion of objects detected among all the objects present in the image. 
Precision (Eq. \ref{eq2:precision}) indicates the proportion of all detected objects that are actually detected. 
The mAP (mean Average Precision) corresponds to a measure of object detection performance based on an IoU (Intersection over Union) between the predicted bounding box and that of the reference data \citep{maharjan2022detection}. 
To this end, the $\text{mAP}_{@50}$ (Eq. \ref{eq3:map50}) uses a 50\% IoU threshold to check whether the detected objects overlap at least 50\% with the real object detected manually.
The $mAP_{[.50:.95]}$ (Eq. \ref{eq4:map5095}) uses a more precise IoU (Eq. \ref{eq5:IOU}) threshold and performs an average accuracy at each overlap threshold over a threshold step of 0.05.
Recall, Precision, $\text{mAP}_{@50}$, $mAP_{[.50:.95]}$ and IoU can be defined as follows

\begin{equation}
\text{Recall} = \frac{\text{True Positives}}{\text{True Positives} + \text{False Negatives}}
\label{eq1:recall}
\end{equation}

\begin{equation}
\text{Precision} = \frac{\text{True Positives}}{\text{True Positives} + \text{False Positives}}
\label{eq2:precision}
\end{equation}

\begin{equation}
    \text{mAP}_{@50} = \frac{1}{N} \sum_{i=1}^{N} \text{AP}_i^{\text{IoU}=0.50}
    \label{eq3:map50}
\end{equation}

where $N$ is the number of classes, $AP_i$ is the average precision for class $i$.

\begin{equation}
    \text{mAP}_{[.50:.95]} = \frac{1}{10} \sum_{k=0}^{9} \left( \frac{1}{N} \sum_{i=1}^{N} \text{AP}_i^{\text{IoU}=0.50+0.05k} \right)
    \label{eq4:map5095}
\end{equation}

where $\text{AP}_i^{\text{IoU}=t}$ is the AP for class $i$ at IoU threshold $t$, $t$ in $\{0.50, 0.55, \dots, 0.95\}$, $N$ is the number of classes.

\begin{equation}
    \text{IoU} = \frac{|\text{B}_{\text{pred}} \cap \text{B}_{\text{gt}}|}{|\text{B}_{\text{pred}} \cup \text{B}_{\text{gt}}|}
    \label{eq5:IOU}
\end{equation}

where \( \text{B}_{\text{pred}} \) is the predicted bounding box and \( \text{B}_{\text{gt}} \) is the ground truth bounding box.

Alongside these performance metrics, indices related to model processing time were measured. 
For each model, the inference time as well as the number of images processed per second on CPU were taken into account \citep{tata2021robotic}.

\subsection{Object size estimation by geometric model}

We used tools from projective geometry \citep{birchfield1998introduction} to extrapolate object dimensions from the bounding boxes provided by the YOLO models.
Estimating real-world sizes from 2D detections requires converting pixel coordinates into 3D features, such as lengths, coordinates, directions, etc.
For this, assumptions about the mapping must be made. We adopt the pinhole camera model, which treats each pixel as corresponding to a ray passing through the camera’s optical center.
This provides a simplified yet effective approximation of the image formation process, ignoring lens distortion and treating the projection as perfectly linear.

We explicitly reconstruct these rays using known intrinsic camera parameters (focal length, sensor size) and extrinsic installation data (camera height and angle), rather than relying on matrix-based projective transformations.
The rays are then intersected with a reference ground plane to recover metric object dimensions, such as width and height, from the projected bounding box edges.

Our framework does not account for real-world optical artifacts such as lens distortion, remaining strictly within the assumptions of the ideal pinhole camera model.
In section~\ref{sec:corrections}, we evaluate the accuracy of our approach by comparing its predicted object dimensions against manually measured ground truth.
While more advanced lens correction techniques exist, they were deemed unnecessary in our context due to the acceptable level of accuracy achieved; however, they could be integrated into the model in future work to further reduce systematic errors~\citep{mohr1996projective}.

We treat each pixel as a ray that originates at the camera’s optical center and passes through the corresponding point on the image sensor, to determine the 3D orientation of a given pixel in the image,

The first step is to compute the effective focal lengths along the horizontal and vertical axes, denoted \( f_x \) and \( f_y \), expressed in \emph{pixels}. These values indicate how many pixels correspond to the camera’s optical focal length in each direction:

\begin{align}
f_x &= \frac{W \cdot f}{w_s} \\
f_y &= \frac{H \cdot f}{h_s}
\label{eq:focal_pixels}
\end{align}

Here, \( f \) is the optical focal length of the camera, in millimeters; \( w_s \) and \( h_s \) are the physical width and height of the sensor, also in millimeters; \( W \) and \( H \) are the image width and height in pixels. The resulting \( f_x \) and \( f_y \) represent the focal length in pixels, i.e., the number of pixels that correspond to one focal length unit along each axis.

Thanks to the focal length, we can reason in pixel coordinates within the image (Figure~\ref{fig6c:screen_box}).
With \( f_x \), \( f_y \), and the image center \( (c_x, c_y) \) defined as:
\[
c_x = \frac{W - 1}{2}, \quad c_y = \frac{H - 1}{2},
\]
we compute the local ray direction associated with any pixel \( (x_{\text{pix}}, y_{\text{pix}}) \) using the following formula:

\begin{align}
\vec{v}_{\text{local}} = \text{normalise}\left(
\begin{bmatrix}
(x_{\text{pix}} - c_x) / f_x \\
(c_y - y_{\text{pix}}) / f_y \\
-1
\end{bmatrix}
\right)
\label{eq:ray_local}
\end{align}

This vector is expressed in the camera’s local coordinate system, where the optical axis is aligned with the negative \( z \)-axis.
The horizontal and vertical terms map the pixel’s displacement from the image center \( (c_x, c_y) \), scaled by the respective focal lengths \( f_x \) and \( f_y \), so that they correspond to physical angular offsets.
Here, the \( y \) coordinates are flipped to match the standard computer vision convention, where the top of the image corresponds to positive \( y \) in pixel space but to negative \( y \) in the 3D camera frame.
These expressions effectively project the focal cone onto a virtual image plane placed at \( z = -1 \): a pixel exactly at the center maps to the direction \( (0, 0, -1) \), while other pixels yield rays tilted accordingly.
The \emph{normalise} operation ensures the resulting vector is a unit-length direction vector.

Finally, the local ray is rotated according to the camera’s orientation.
In general, a camera’s orientation is described by three successive rotations: a pitch angle \( \phi \) (rotation around the horizontal \( x \)-axis), a yaw angle \( \theta \) (rotation around the vertical \( y \)-axis), and a roll angle (rotation around the optical \( z \)-axis).
Each of these affects the perceived direction of rays in different ways.

In our setup, only the pitch angle \( \phi \), which controls the vertical inclination of the camera, is taken into account.
The yaw angle \( \theta \) is ignored, as it rotates the camera parallel to the ground and does not affect object size estimation.
The roll angle remains zero throughout the acquisition, as the camera stays horizontally aligned.

We can, therefore, define a rotation matrix as

\begin{align}
R_x(\phi) =
\begin{bmatrix}
1 & 0 & 0 \\
0 & \cos\phi & -\sin\phi \\
0 & \sin\phi & \cos\phi
\end{bmatrix}
\end{align}

and compute the ray direction in actual world coordinates:

\begin{align}
\vec{v}_{\text{world}} = R_x(\phi) \cdot \vec{v}_{\text{local}},
\label{eq:ray_world}
\end{align}

where $\vec{v}_{\text{local}}$ is the pixel direction computed from equation~\ref{eq:ray_local}.

This mapping from image pixels to 3D rays forms the geometric backbone of our estimation process: from these rays, we construct planes, compute intersections, and measure distances to recover metric dimensions.
For conciseness, we omit the derivation of these standard geometric operations (plane construction, ray-plane intersections, point-to-plane distances); they are recapped in the appendix for reference.

We compute five directional vectors of interest: \( \vec{v}_{\text{top-left}} \), \( \vec{v}_{\text{top-right}} \), \( \vec{v}_{\text{bottom-left}} \), \( \vec{v}_{\text{bottom-right}} \), and \( \vec{v}_{\text{center}} \).
Each vector corresponds to a key point of the bounding box: its four corners and its center.
This process is summarised in Figure~\ref{fig6d:projected_vectors}.
They indicate the direction in world space along which that point is observed from the camera.

From the four corner vectors, we construct four planes by pairing each adjacent corner with the camera’s optical center, denoted as $\vec{C}$. For example, the top edge of the bounding box defines the plane:

\[
\Pi_{\text{top}} = \text{Plane}(\vec{C}, \vec{v}_{\text{top-left}}, \vec{v}_{\text{top-right}}),
\]

and similarly for the bottom, left, and right edges. These planes form a pyramid-like volume spanning from $\vec{C}$ that encloses the region in space where the object is assumed to lie.

Since the object’s distance from the camera is not known \textit{a priori}, we estimate its position by intersecting the central ray \( \vec{v}_{\text{center}} \) with the ground plane (Fig.~\ref{fig6e:projected_object}).
This intersection yields a single 3D point, denoted as $\vec{O}$ that we take as a proxy for the object’s center location in space.

The object’s width and height are then estimated by summing the distances from $\vec{O}$ to the pairs of vertical and horizontal bounding planes:

\begin{align}
dim_x &= \text{distance}(\vec{O}, \Pi_{\text{left}}) + \text{distance}(\vec{O}, \Pi_{\text{right}}) \\
dim_y &= \text{distance}(\vec{O}, \Pi_{\text{top}}) + \text{distance}(\vec{O}, \Pi_{\text{bottom}}).
\label{eq:size_estimation}
\end{align}

Figure~\ref{fig6f:point_plane_distance} shows a visual interpretation of these distances from the camera's perspective.

This geometric pipeline provides a fully interpretable method for estimating the object dimensions $dim_x$ and $dim_y$ from monocular images. However, due to real-world deviations from the ideal pinhole model—such as lens distortion, imperfect camera alignment, wrong bounding box positioning, and ground unevenness—we observe slight discrepancies between predicted and actual object sizes. In the next section, we describe how an \textit{a posteriori} correction is applied to compensate for these systematic biases and improve the accuracy of our estimates. Formulas are detailed in the Appendix A.

\begin{figure}[H]
    \centering

    \begin{subfigure}[t]{0.49\textwidth}
        \centering
        \includegraphics[width=\textwidth]{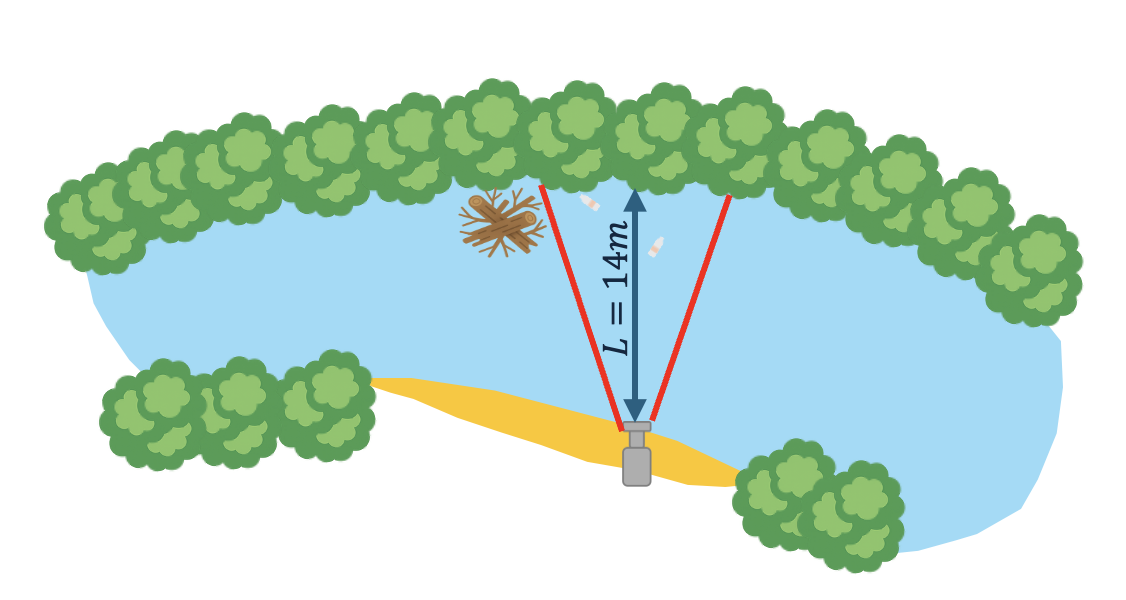}
        \caption{Camera position according to the watercourse.}
        \label{fig6a:cam_watercourse_position}
    \end{subfigure}
    \hfill
    \begin{subfigure}[t]{0.49\textwidth}
        \centering
        \includegraphics[width=\textwidth]{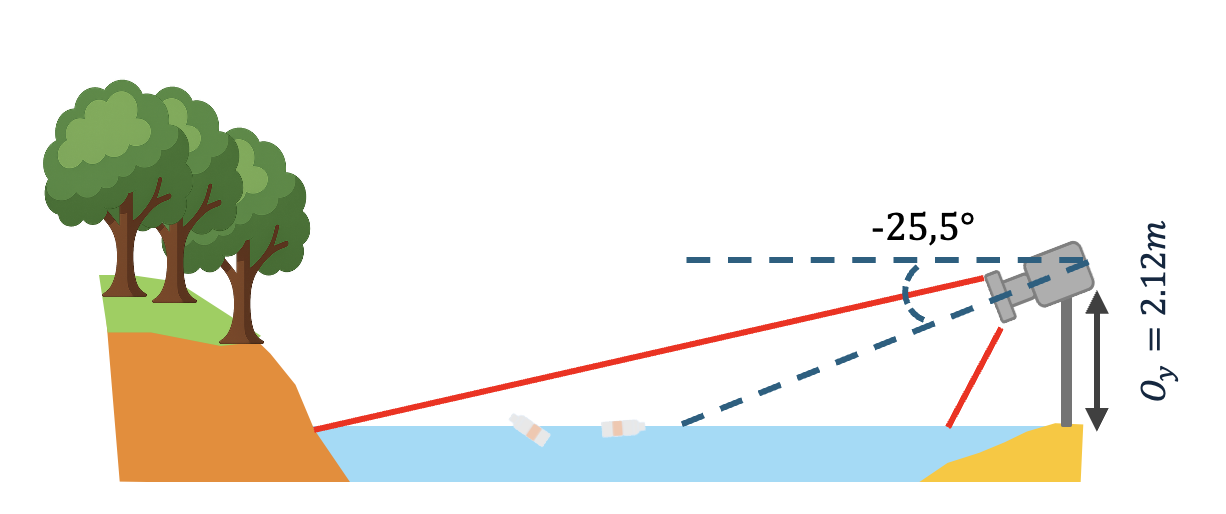}
        \caption{Camera angle.}
        \label{fig6b:cam_watercourse_angle}
    \end{subfigure}

    \vspace{1em}

    \begin{subfigure}[t]{0.33\textwidth}
        \centering
        \resizebox{\textwidth}{!}{\input{figure/method/fig_screen_box.tikz.tex}}
        \caption{Screen box definition.}
        \label{fig6c:screen_box}
    \end{subfigure}
    \hfill
    \begin{subfigure}[t]{0.63\textwidth}
        \centering
        \resizebox{\textwidth}{!}{\input{figure/method/fig_projected_vectors.tikz.tex}}
        \caption{Direction vectors.}
        \label{fig6d:projected_vectors}
    \end{subfigure}

    \vspace{1em}

    \begin{subfigure}[t]{0.49\textwidth}
        \centering
        \resizebox{\textwidth}{!}{\input{figure/method/fig_projected_object.tikz.tex}}
        \caption{Bounding box's center projected onto the ground plane.}
        \label{fig6e:projected_object}
    \end{subfigure}
    \hfill
    \begin{subfigure}[t]{0.49\textwidth}
        \centering
        \resizebox{\textwidth}{!}{\input{figure/method/fig_point_plane_distance.tikz.tex}}
        \caption{Distance from $O$ to the planes.}
        \label{fig6f:point_plane_distance}
    \end{subfigure}

    \caption{
        Visualisation of the object size estimation process.
        Subfigures~\ref{fig6a:cam_watercourse_position} and~\ref{fig6b:cam_watercourse_angle} illustrate the camera's field of view setup.
        Subfigure~\ref{fig6c:screen_box} shows the bounding box in 2D. 
        Subfigure~\ref{fig6d:projected_vectors} illustrates how this box is lifted into 3D space via ray projections from the camera center.
        Subfigure~\ref{fig6e:projected_object} depicts the intersection of these rays with the ground plane, forming a spatial footprint of the object.
        Finally, Subfigure~\ref{fig6f:point_plane_distance} shows the 4 distances from the camera's perspective.
    }
    \label{fig6:screen_box_pipeline}
\end{figure}

\subsection{Metrics and size of detected bounding box correction}\label{sec:corrections}

A pipeline of two corrections was applied to the dimensions predicted by the geometric model on the bounding boxes predicted by the object detection model.

The first correction (Bounding box shape correction) corrects the shapes of the bounding boxes predicted by the AI model.
This correction uses the predicted dimensions of the bounding boxes detected by the AI model as the data to be corrected.
The reference data corresponds to the dimensions predicted on the manually annotated bounding boxes (Fig.\ref{fig7:corr_pip}).
Here, the aim is to correct systematic biases in the estimated physical dimensions of detected objects, caused by imperfect alignment between the annotated bounding boxes and the actual physical extents of the objects.
One must note that this correction is applied solely in the physical space, without relying on the bounding box coordinates themselves.

The second correction (Predicted dimensions correction) aims to correct the dimensions (width and length) predicted by the geometric model on the corrected shape of the bounding boxes (Fig.\ref{fig7:corr_pip}).
Here, the reference data is created by manually measuring the dimensions of the objects.
Refinement of predicted dimensions is sought with this correction. 

As a result, we expect each individual correction to improve the precision of the method. We will also look at the relevance of combining the two corrections

\begin{figure}[H]
    \centering
    \includegraphics[width=1\linewidth]{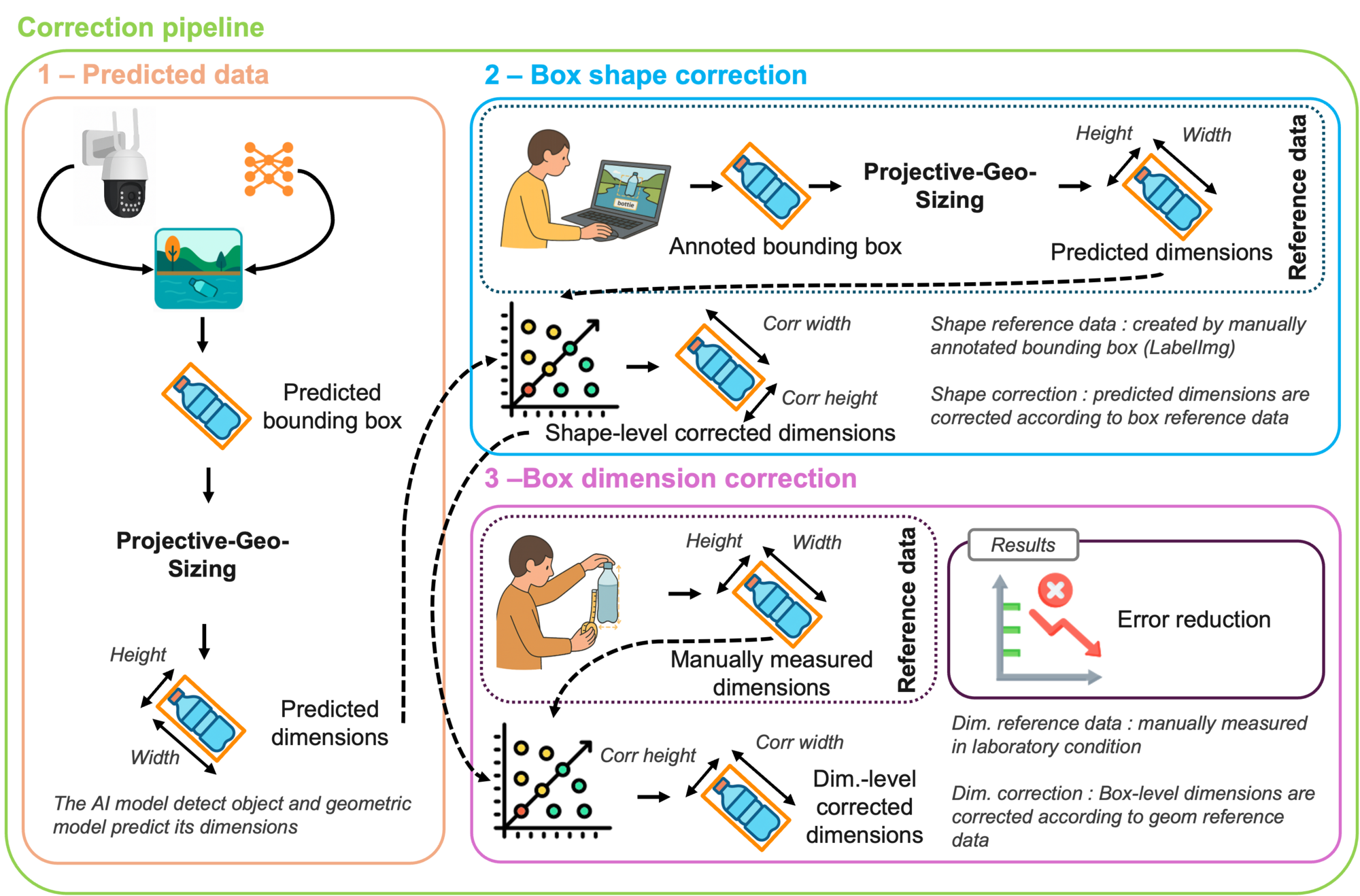}
    \caption{Regression-based correction pipeline}
    \label{fig7:corr_pip}
\end{figure}

\subsubsection{Regression-based correction}

For both corrections, two parametric regression models were used : simple linear regression and polynomial regression. 
In parametric regression, the number of parameters is finite and a specific functional form is assumed in advance to describe the relationship between variables \citep{mahmoud2019parametric}.

\subsubsection{Performance metrics}

For each regression, the correlation coefficient is extracted. In order to assess the performance of the latter, the root mean square error (RMSE) and the mean absolute error (MAE) are calculated before and after the correction. 
The RMSE is defined by : 

\begin{align}
\text{RMSE} = \sqrt{ \frac{1}{n} \sum_{i=1}^{n} (y_i - \hat{y}_i)^2 }
\label{eq:RMSE}
\end{align}

and the MAE is calculated by : 

\begin{align}
\text{MAE} = \frac{1}{n} \sum_{i=1}^{n} \left| y_i - \hat{y}_i \right|
\label{eq:MAE}
\end{align}

\subsubsection{Size-estimation validation protocol}

The quantitative evaluation of the size-estimation pipeline was performed using twenty objects from the Steingiessen dataset, each measured manually. 
The evaluated objects correspond to anthropogenic debris introduced during the controlled acquisition protocol, including items such as plastic bottles, containers, and other floating objects with relatively simple geometries.

The ground-truth dimensions were measured manually, with object widths ranging from 20.5 cm to 33.5 cm and heights from 5.5 cm to 10 cm (Tab. \ref{tab:size_detailed} Appendix B). 
These objects represent a limited range of shapes, primarily compact or moderately elongated geometries, and do not fully capture the diversity of morphologies encountered in natural river systems.

The evaluated instances were observed within the effective field of view of the bank-mounted camera, corresponding to a restricted range of distances from the camera (from 1 to 14 meters).
In addition, all measurements were acquired under relatively stable low flow conditions, with no significant variation in water level during the experiment ($\sim 1.8$ meters).

As a result, the validation dataset represents a constrained subset of possible observation conditions, and should be interpreted as a controlled experimental domain rather than a comprehensive representation of real-world riverine debris.

\begin{figure}[H]
    \centering
    \includegraphics[width=1\linewidth]{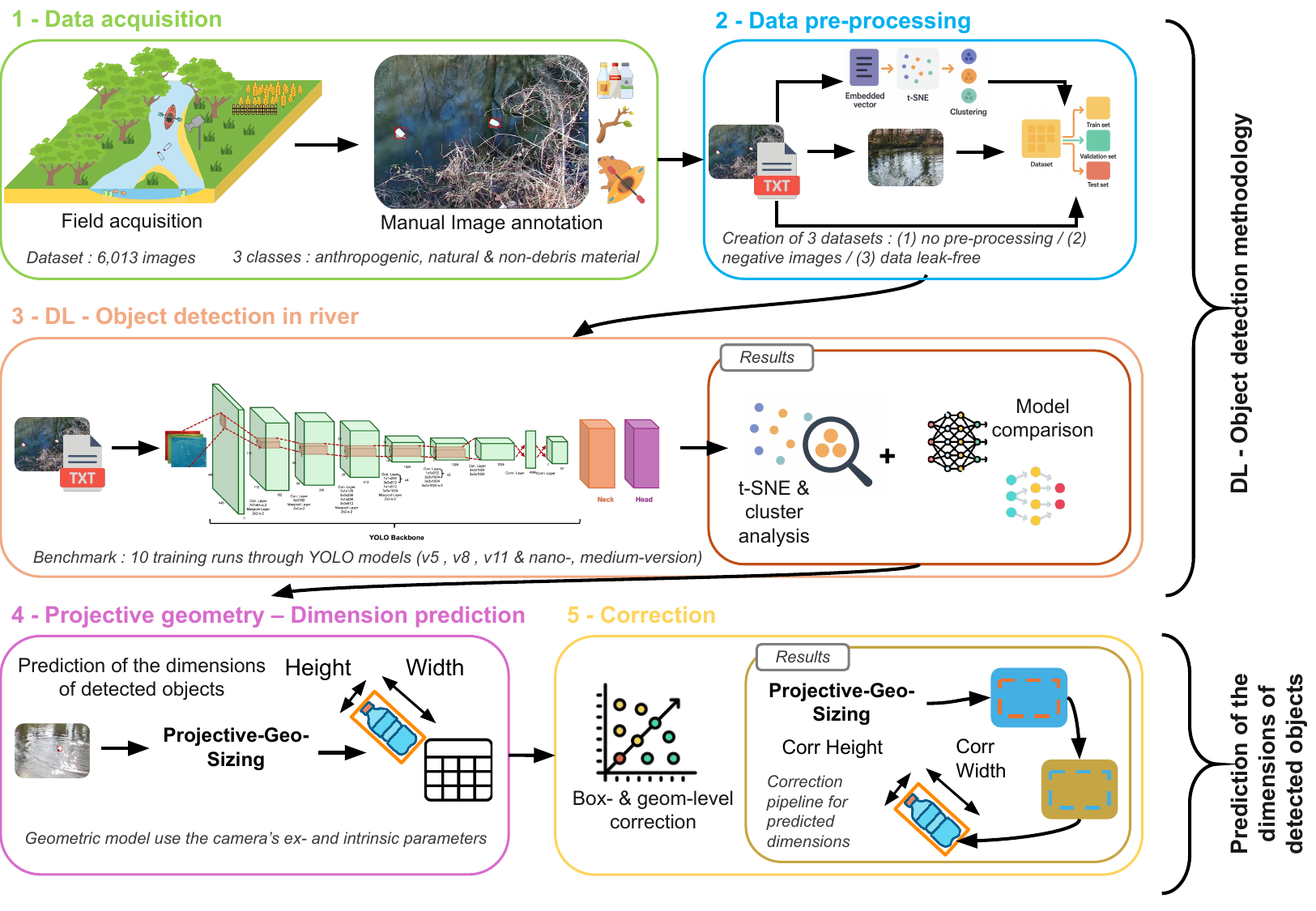}
    \caption{Employed methodology : (1) Acquired data is manually annotated to constitute initial dataset. (2) The latter is pre-processed so that (3) several models can then be trained to detect debris in rivers. (4) Dimensions of detected objects are predicted and then (5) corrected to reduce errors}
    \label{fig1:employed_methodology}
\end{figure}

\section{Experimental results}

\subsection{Leak-free dataset construction using t-SNE and DBSCAN}

Subsequent to the dimensional reduction of the data, the DBSCAN clustering methodology was implemented to avoid time leakage and overestimation of the performance model during the training of the data.
The t-SNE reveals a structured organization with localized concentrations of points (Fig.\ref{fig8:t-SNE}). 
These aggregations of points serve to emphasize images that exhibit analogous characteristics.
Furthermore, within the reduced space, point aggregations are distinguished from others by the absence of apparent continuity.

Applying the DBSCAN algorithm to the dimensionally reduced data resulted in 53 distinct clusters.
The DBCV (Density-Based Clustering Validation) score of 0.98 indicates highly compact and well-separated clusters.
Visual inspection of clusters 33, 45, and 52 highlights two distinct phenomena: a clear separation between upstream and downstream camera scenes, and a temporal separation of scenes from the same camera based on the presence of anthropogenic debris (Fig.\ref{fig8:t-SNE}).
Cluster 33 includes only upstream images, while cluster 45 and 52 contain downstream scenes, each showing different debris compositions.
This suggests that images from the same scene can be found in the same data set (Fig.\ref{fig8:t-SNE}) rather than several data sets; thus reducing the risk of temporal data leakage.

As previously stated, clusters were separated into three groups : train, validation, and test, with the following proportions : 80\%, 10\%, 10\% respectively. 
At first, the t-SNE reveals that no cluster was separated into two different groups, which means that all images from the same cluster are actually in the same sub-dataset (Fig.\ref{fig8:t-SNE}). 

\begin{figure}[H]
    \centering
    \includegraphics[width=1\linewidth]{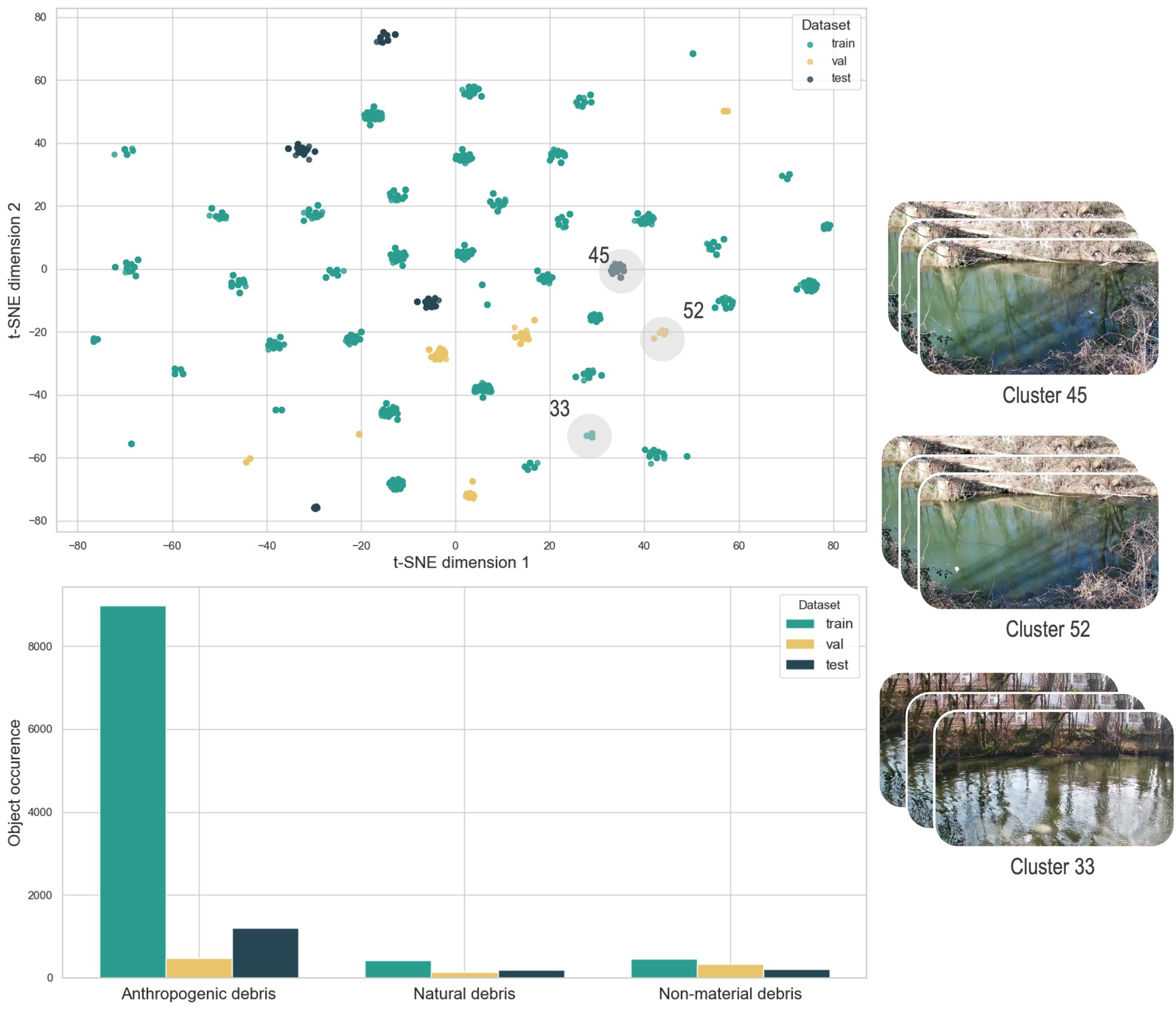}
    \caption{Clusters applied to dimensionally reduced data, occurrence of object classes in each dataset (Train, Validation, Test) and visual examples of clusters 33, 45 and 52}
    \label{fig8:t-SNE}
\end{figure}

\subsection{Detection performance and dataset benchmark effects}

In this section, we first evaluated the baseline performance of the tested YOLO architectures. It then analyses the effect of various dataset design choices, such as temporal leakage and the proportion of negative images and class weighting on detection performance.

\subsubsection{Baseline benchmark of YOLO architectures}

We first evaluate six configurations on the baseline dataset with a random split. These configurations included the YOLOv5, YOLOv8 and YOLOv11 models, as well as their nano and medium variants.

Among trainings 1 to 6, with an initial examination of the technical characteristics (Tab.\ref{tab1:training_model}) shows that Train 6 (YOLOv11-m) has the longest training time (21.508 hours for 134 epochs) while Train 1 (YOLOv5-n) is the fastest (2.085 hours for 80 epochs).
With regard to time inference performance, Train 1 presents a favourable trade-off between performance and computational cost (135.27 ms for 7.39 FPS) on CPU, while Train 6 is the least efficient (537.08 ms for 1.86 FPS).
Concerning performance metrics (Tab.\ref{tab2:metric_model}), Train 4 (YOLOv8-m) and Train 6, have the best mAP@50 (0.965) and mAP@50-95 (0.747) respectively. 
By contrast, Train 1 has the lowest mAP@50 (0.933) and mAP@50-95 (0.655). 
This finding suggests that Trains 4 and 6 perform well in detecting and localising objects in our experimental setup, while Train 1 shows the greatest difficulty.

Furthermore, Train 4 and Train 3 (YOLOv8-n) have the best recall (0.942) and precision (0.97), respectively, while Train 1 has the worst recall (0.885) and Train 5 (YOLOv11-m) the worst precision (0.925).

Among the six trained models, two are kept for further analysis : Train 3 and Train 6.
This selection is based on a trade-off between detection performance and CPU inference time, considering future deployment on devices like Raspberry Pi.
Train 1, despite its high inference speed (135.27 ms for 7.39 FPS), shows limited detection performance.
Train 2 (YOLOv5-m) obtains good detection results, but is too slow (466.17 ms for 2.15 FPS).
Train 4 achieves excellent precision (0.954) and recall (0.942), but its inference time (517.54 ms for 1.93 FPS) and the marginal gain of mAP@50 (0.965 compared with 0.962 for train 6) do not justify its selection.
While Train 5 demonstrates strong detection capabilities, its performance remains inferior to Train 6. 
Furthermore, its precision (0.925) is lower than that of Train 3 (0.970), although it achieves a faster inference speed (137.10 ms compared to 143.12 ms).

Finally, Train 6 is selected for its strong overall metrics (Tab.\ref{tab2:metric_model}).
Though slower (537.08 ms for 1.86 FPS), it is well-suited for robust, out-of-board detection.
Train 3 is chosen as a balanced option for embedded applications, with good metrics (Tab.\ref{tab2:metric_model}) and efficient inference (Tab.\ref{tab1:training_model}).

\begin{table}[H]
\centering
\scriptsize

\begin{tabular}{clcccc}
\hline
 \textbf{Train}&\textbf{Model} & \textbf{Epoch} & \textbf{Proc. time (h)} & \textbf{Inf. time CPU (ms)}& \textbf{FPS CPU}\\
\hline
 1&yolov5n    & 80  & 2.085  & \textbf{135.27}& \textbf{7.39}\\
 2&yolov5m    & 104 & 11.234 & 466.17 & 2.15  \\
 3&yolov8n    & 87  & 2.332  & 143.12 & 6.99  \\
 4&yolov8m    & 112 & 14.537 & 517.54 & 1.93  \\
 5&yolov11n   & 104 & 3.312  & 137.10 & 7.29  \\
 6&yolov11m   & 134 & 21.508 & 537.08 & 1.86  \\
\hline
\end{tabular}
\caption{Technical drive characteristics of models}
\label{tab1:training_model}
\end{table}

\begin{table}[H]
\centering
\scriptsize

\begin{tabular}{clcccc}
\hline
 \textbf{Train}&\textbf{Model} & \textbf{mAP50} & \textbf{mAP50-95} & \textbf{Recall} & \textbf{Precision} \\
\hline
 1&yolov5n    & 0.933 & 0.655 & 0.885 & 0.936 \\
 2&yolov5m    & 0.951 & 0.708 & 0.908 & 0.949 \\
 3&yolov8n    & 0.939 & 0.672 & 0.888 & \textbf{0.970}\\
 4&yolov8m    & \textbf{0.965}& 0.728 & \textbf{0.942}& 0.954 \\
 5&yolov11n   & 0.946 & 0.694 & 0.912 & 0.925 \\
 6&yolov11m   & 0.962 & \textbf{0.747} & 0.937 & 0.955 \\
\hline
\end{tabular}
\caption{Model performance metrics}
\label{tab2:metric_model}
\end{table}

\subsubsection{Effect of temporal leakage on detection performance}

We evaluate the impact of temporal leakage by comparing models trained on the initial randomly split dataset with models trained on the leak-free dataset constructed using the t-SNE \& DBSCAN-based clustering strategy.
Training the models with a dataset without data leakage addresses the problem of over- or under-interpretation of their performance metrics.

Training the YOLOv8-n model without data leakage readjusts its performance metrics significantly (Fig.\ref{fig9:yolov8n_results_new}).
While the true positive (TP) rate for class 3 remains stable (1.00 in both configurations), the rates for classes 1 and 2 significantly decrease, dropping from 0.97 to 0.63 and from 0.82 to 0.29, respectively.
In the leak-free setting, false negatives (FN) increase substantially: 37\% of class 1 instances and 56\% of class 2 instances are missed and predicted as background, compared to only 3\% and 18\% in the baseline.
Additionally, the column background reveals increased false positive (FP) rates, with 0.65 and 0.06 spurious detections for classes 1 and 2 respectively, indicating that the model also produces erroneous detections in empty regions.
This downward trend is reflected in the global metrics, where precision drops from 0.970 to 0.551, recall from 0.888 to 0.508, and mAP@50 from 0.939 to 0.488 (Fig.\ref{fig9:yolov8n_results_new}).
The mAP@50-95 is similarly affected, falling from 0.672 to 0.261, highlights the impact of leakage across all IoU thresholds.

Training the YOLOv11-m model without data leakage produces consistent results (Fig.\ref{fig11:yolov11m_results_new}).
While the TP rate for class 3 remains unaffected (1.00), the rates for classes 1 and 2 decrease from 0.99 to 0.70 and from 0.92 to 0.60, respectively.
FN also increase markedly, with 30\% and 37\% of class 1 and class 2 instances respectively predicted as background, compared to only 1\% and 8\% in the baseline.
These trends are reflected in the global metrics: precision drops from 0.955 to 0.491, recall from 0.937 to 0.529, and mAP@50 from 0.962 to 0.536 (Fig.\ref{fig11:yolov11m_results_new}).
The mAP@50-95 decreases from 0.747 to 0.306, confirming the substantial inflation of performance metrics caused by temporal leakage.

Comparing both architectures in the leak-free setting, YOLOv11-m consistently outperforms YOLOv8-n across all metrics. 
Precision is higher (0.491 vs. 0.551, in favour of YOLOv8-n for precision alone, but 0.536 vs. 0.488 for mAP@50), and the TP rates for classes 1 and 2 are better preserved in YOLOv11-m (0.70 and 0.60) compared to YOLOv8-n (0.63 and 0.29).
This suggests that the larger YOLOv11-m architecture generalises better to unseen temporal sequences, even when trained without data leakage.
Overall, training without data leakage has enabled the metrics to be adjusted to reflect the actual performance of the models, and confirms that YOLOv11-m is the more robust architecture for this task.

\begin{figure}[H]
    \centering
    \includegraphics[width=1\linewidth]{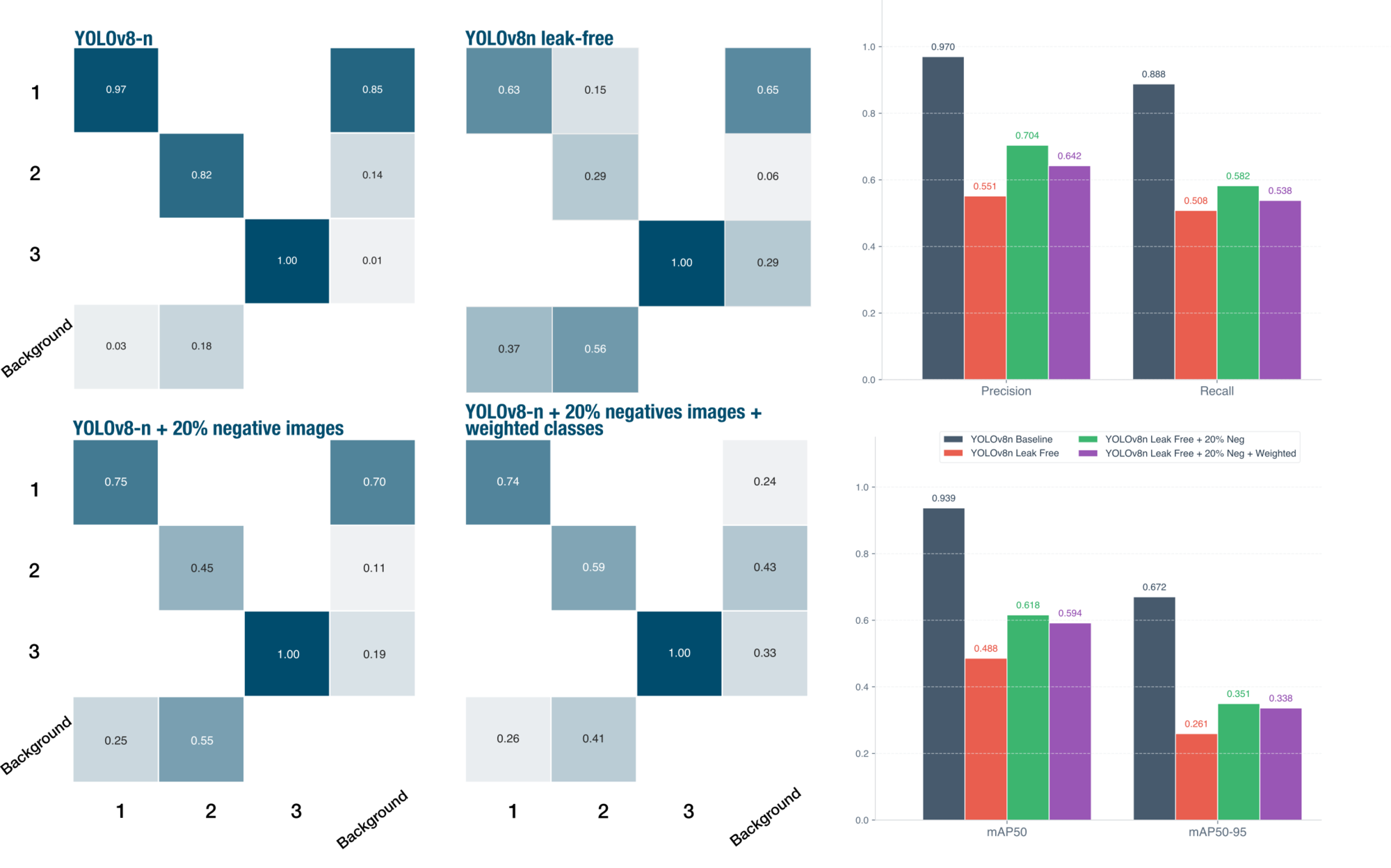}
    \caption{YOLOv8-n training result where 1 = Anthropogenic Debris ; 2 = Natural Debris ; 3 = Non-Debris Material}
    \label{fig9:yolov8n_results_new}
\end{figure}

\subsubsection{Effect of negative-image proportion on detection performance}

Five different proportions of negative images, ranging from 0\% to 40\%, are evaluated using both YOLOv8-n and YOLOv11-m to determine the optimal balance for training (Fig.\ref{fig10:ratio_image_neg}).
For both architectures, recall and mAP@50 reach their peak at 20\%, with YOLOv11-m achieving a mAP@50 of 0.648 and a recall of 0.630, and YOLOv8-n reaching 0.618 and 0.582 respectively.
Beyond this threshold, both metrics decline steadily, which suggests that an excess of negative images shifts the model towards over-conservatism, causing it to miss true detections.
Regarding FP, the total count is minimised around 20\% for YOLOv11-m (dropping to approximately 360), while YOLOv8-n also reaches its lowest FP count at this proportion (approximately 340).
At higher proportions (30\% and 40\%), FP counts rise again for both models, indicating that the benefit of negative images plateaus and eventually reverses.
Based on this analysis, 20\% is selected as the optimal proportion of negative images for all subsequent experiments, as it provides the best trade-off between detection performance and FP reduction for both architectures.

\begin{figure}[H]
    \centering
    \includegraphics[width=1\linewidth]{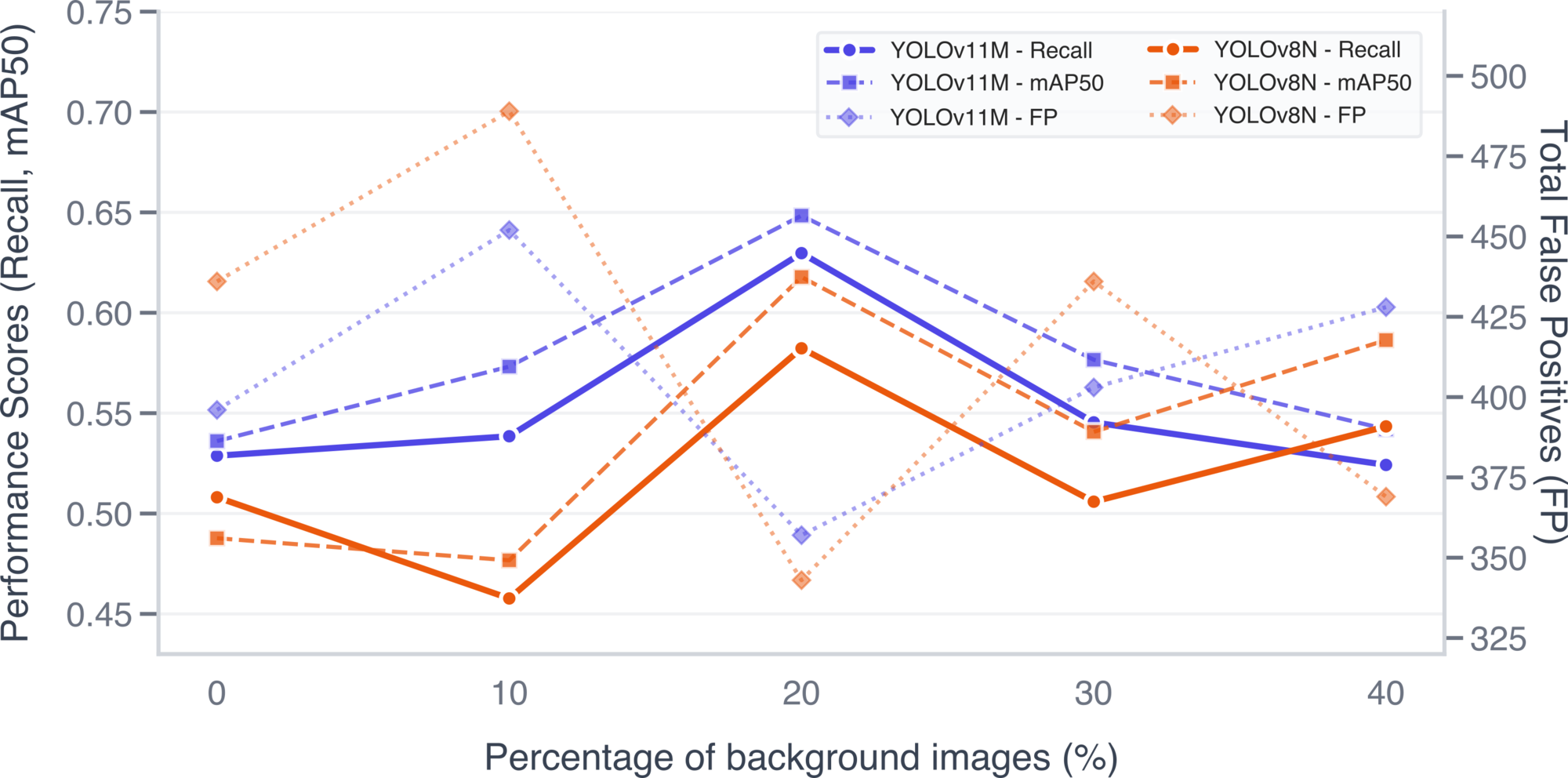}
    \caption{Effects of negative images proportion in dataset on YOLOv8-n \& YOLOv11-m training perofmance}
    \label{fig10:ratio_image_neg}
\end{figure}

In this section, the training process with the YOLOv8-n and YOLOv11-m models therefore uses a dataset comprising the leak-free dataset augmented with 20\% negative images. 
These trainings address the hypothesis that negative images reduce false positives in the model's detections.

The results for the YOLOv8-n model show a notable improvement compared to the leak-free baseline across all evaluation metrics (Fig.\ref{fig9:yolov8n_results_new}).
In the confusion matrices, class 1 (anthropogenic debris) recovers a TP rate of 0.75, up from 0.63 in the leak-free configuration.
Class 2 (natural debris) also improves, rising from 0.29 to 0.45.
Class 3 (non-debris materials) remains perfectly detected (1.00) across all configurations.
The FN rate decreases for class 1 (from 0.37 to 0.25), though class 2 remains difficult, with 55\% of instances still predicted as background.
These gains are confirmed by the global metrics: precision increases from 0.551 to 0.704, recall from 0.508 to 0.582, and mAP@50 from 0.488 to 0.618 (Fig.\ref{fig9:yolov8n_results_new}).
The mAP@50-95 similarly improves from 0.261 to 0.351, indicating more spatially precise detections.

The results for the YOLOv11-m model follow the same trend (Fig.\ref{fig11:yolov11m_results_new}).
Class 1 recovers a TP rate of 0.77 (up from 0.70 in the leak-free setting) but class 2 decreases from 0.60 to 0.48.
Class 3 remains stable at 1.00.
FN for class 1 decrease from 0.30 to 0.23 but increase from 0.37 to 0.52 for class 2, suggesting that while negative images help reduce missed detections for class 1, class 2 remains challenging.
Globally, precision rises from 0.491 to 0.704, recall from 0.529 to 0.630, and mAP@50 from 0.536 to 0.648 (Fig.\ref{fig11:yolov11m_results_new}).
The mAP@50-95 increases from 0.306 to 0.402, which confirms the consistent improvement brought by the addition of negative images.

Across both architectures, the inclusion of 20\% negative images consistently improves detection performance over the leak-free baseline, with YOLOv11-m maintaining its advantage over YOLOv8-n in all metrics.
This confirms that negative images help the models better discriminate between true detections and empty background regions, and that this strategy is complementary to the leak-free training protocol.

\begin{figure}[H]
    \centering
    \includegraphics[width=1\linewidth]{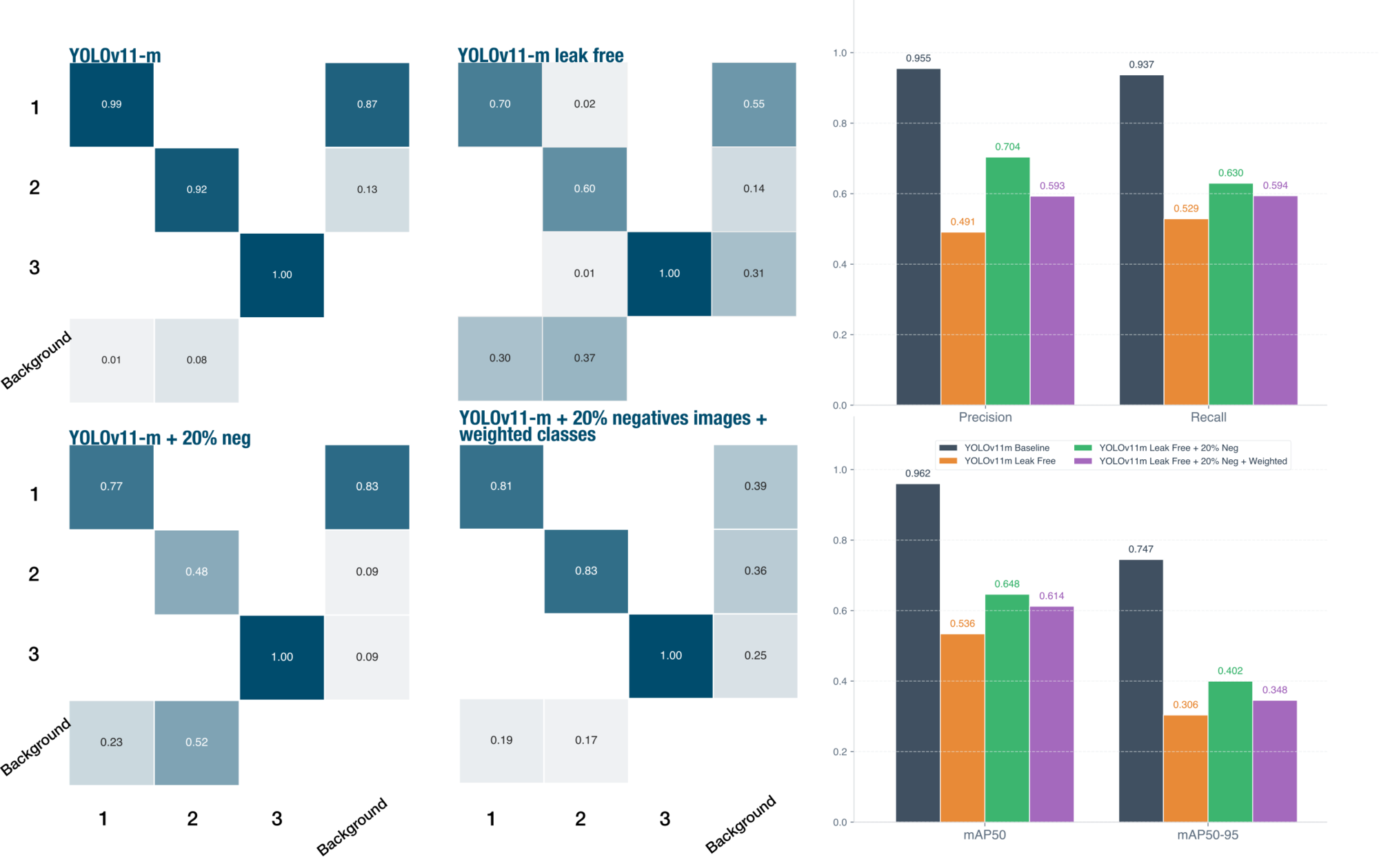}
    \caption{YOLOv11-m training results where 1 = Anthropogenic Debris ; 2 = Natural Debris ; 3 = Non-Debris Material}
    \label{fig11:yolov11m_results_new}
\end{figure}

\subsubsection{Effect of class weighting on detection performance}

In this section, we investigate whether applying class weights during training can further improve detection performance beyond what is achieved with the leak-free dataset and 20\% negative images.

For the YOLOv8-n model, the addition of class weights produces mixed results compared to the 20\% negative images configuration (Fig.\ref{fig9:yolov8n_results_new}).
In the confusion matrices, class 1 maintains a comparable TP rate (0.74 vs. 0.75), and class 3 remains perfectly detected (1.00).
Class 2 shows a meaningful improvement, with its TP rate rising from 0.45 to 0.59, suggesting that class weighting effectively helps the model focus more attention on this difficult class.
However, although the FP rate in the background column decreases noticeably for class 1 (from 0.70 to 0.24 — a reduction), it increases for class 2 (from 0.11 to 0.43), indicating that the model produces more spurious detections in empty regions when weights are applied.
The FN rate for class 1 decreases slightly (from 0.25 to 0.26), while class 2 FN improve (from 0.55 to 0.41).
Globally, the weighted configuration underperforms slightly relative to the 20\% negative images setting: precision decreases from 0.704 to 0.642, while recall improves marginally from 0.582 to 0.538.
The mAP@50 drops from 0.618 to 0.594, and the mAP@50-95 decreases from 0.351 to 0.338 (Fig.\ref{fig9:yolov8n_results_new}).
This suggests that for YOLOv8-n, class weighting redistributes performance across classes without providing a net global gain, and introduces a precision-recall trade-off that slightly penalises overall detection quality.

For the YOLOv11-m model, the effect of class weighting is similarly ambivalent (Fig.\ref{fig11:yolov11m_results_new}).
Class 1 TP rate improves from 0.77 to 0.81, and class 2 shows a substantial gain, rising from 0.48 to 0.83 — the highest value observed across all leak-free configurations for this class.
Class 3 remains stable at 1.00.
However, this improvement in TP rates is accompanied by a marked increase in FP: the background column shows values of 0.36 for class 2, compared to 0.09 in the 20\% negative images setting, indicating that the model becomes more aggressive in its detections and generates more erroneous predictions in background regions.
FN rates decrease for both class 1 (0.23 to 0.19) and class 2 (0.52 to 0.17), confirming that class weighting successfully reduces missed detections.
At the global level, precision decreases from 0.704 to 0.593, while recall improves from 0.630 to 0.594.
The mAP@50 drops slightly from 0.648 to 0.614, and the mAP@50-95 decreases from 0.402 to 0.348 (Fig.\ref{fig11:yolov11m_results_new}).

Across both architectures, class weighting consistently improves the detection of class 2 (natural debris), which is the most challenging category in this dataset, at the expense of increased false positives 
and a slight reduction in global metrics.
This reflects a fundamental precision-recall trade-off: weighted training pushes the models to detect more instances, including in ambiguous regions, which benefits recall but penalises precision.
YOLOv11-m benefits more from class weighting than YOLOv8-n in terms of per-class TP recovery, particularly for class 2, though both models exhibit the same global degradation pattern.
Overall, the 20\% negative images configuration without class weighting remains the best-performing strategy in terms of global metrics (mAP@50 and mAP@50-95), while class weighting may be preferable in use cases where minimising missed detections for natural debris is prioritised over overall precision.

\subsection{Regression-based correction}
\subsubsection{Predicted bounding box shape correction}
\label{sec:bb_correction}

The application of a correction to the YOLOv11-m model (without data leakage, with 20\% negative image and class weighting) has been demonstrated to enhance the accuracy of the predicted bounding box shape dimensions (Fig.\ref{fig12:regression_correction}a \& b).

In the absence of correction, height predictions are biased, as evidenced by an RMSE of 2.92 cm and a MAE of 1.70 cm (Fig.\ref{fig12:regression_correction}a).
This bias is characterised by a systematic tendency to overestimate, as demonstrated by the residual distribution being skewed towards positive values, with a median of 0.00 cm, a Q1 of 0.00 cm and a Q3 of 1.90 cm, indicating that the majority of predictions exceed the ground truth.
Linear regression partially reduces these errors (RMSE = 1.84 cm; MAE = 1.49 cm; $R^2 = 0.22$; $p = 3.37 \times 10^{-13}$), and shifts the residual distribution towards more negative values (median = --0.21 cm; Q1 = --1.24 cm; Q3 = 1.42 cm), suggesting that the correction overcorrects a portion of predictions.
Polynomial regression ($y = -0.080x^2 + 2.38x - 5.75$) yields the best results for height (RMSE = 1.69 cm; MAE = 1.17 cm; $R^2 = 0.34$; $p = 2.45 \times 10^{-20}$).
The residuals become more centred and symmetrical around zero (median = --0.035 cm; Q1 = --0.840 cm; Q3 = 0.638 cm), with reduced extremes ranging from --2.90 cm to 2.69 cm, compared to --2.70 cm to 4.50 cm without correction.
This confirms that polynomial regression effectively reduces both the magnitude and the asymmetry of height prediction errors.

For width predictions, the uncorrected model already demonstrates reasonable performance (RMSE = 2.79 cm; MAE = 1.46 cm; $R^2 = 0.78$), with residuals distributed around a near-zero median of 0.05 cm (Q1 = 0.00 cm; Q3 = 1.80 cm), indicating a slight tendency to underestimate ground truth widths (Fig.\ref{fig12:regression_correction}b).
Linear regression ($y = 0.88x + 1.29$) partially improves the fit ($R^2 = 0.81$; $p = 5.73 \times 10^{-51}$), reducing RMSE to 2.56 cm but increasing MAE to 1.52 cm. 
This approach also recentres the residuals (median = 0.13 cm; Q1 = --1.14 cm; Q3 = 1.24 cm) and narrowing the error range from 
--3.83 cm to 4.31 cm.
Polynomial regression ($y = 0.008x^2 + 0.61x + 3.36$) achieves comparable performance ($R^2 = 0.82$; $p = 3.08 \times 10^{-80}$; RMSE = 2.54 cm; MAE = 1.61 cm), with a similar residual distribution (median = 0.16 cm; Q1 = --1.38 cm; Q3 = 1.36 cm) and comparable 
extremes (--4.08 cm to 4.38 cm).
The marginal difference between the two regression methods suggests that the relationship between YOLO-predicted and ground truth widths is largely linear, and that the added complexity of the polynomial term provides no meaningful additional benefit for width correction.
Overall, both regression approaches yield consistent improvements for width, primarily by reducing the positive bias present in the uncorrected predictions and tightening the interquartile range of residuals.

\begin{figure}[H] 
    \centering
    \includegraphics[width=0.8\textwidth, keepaspectratio]{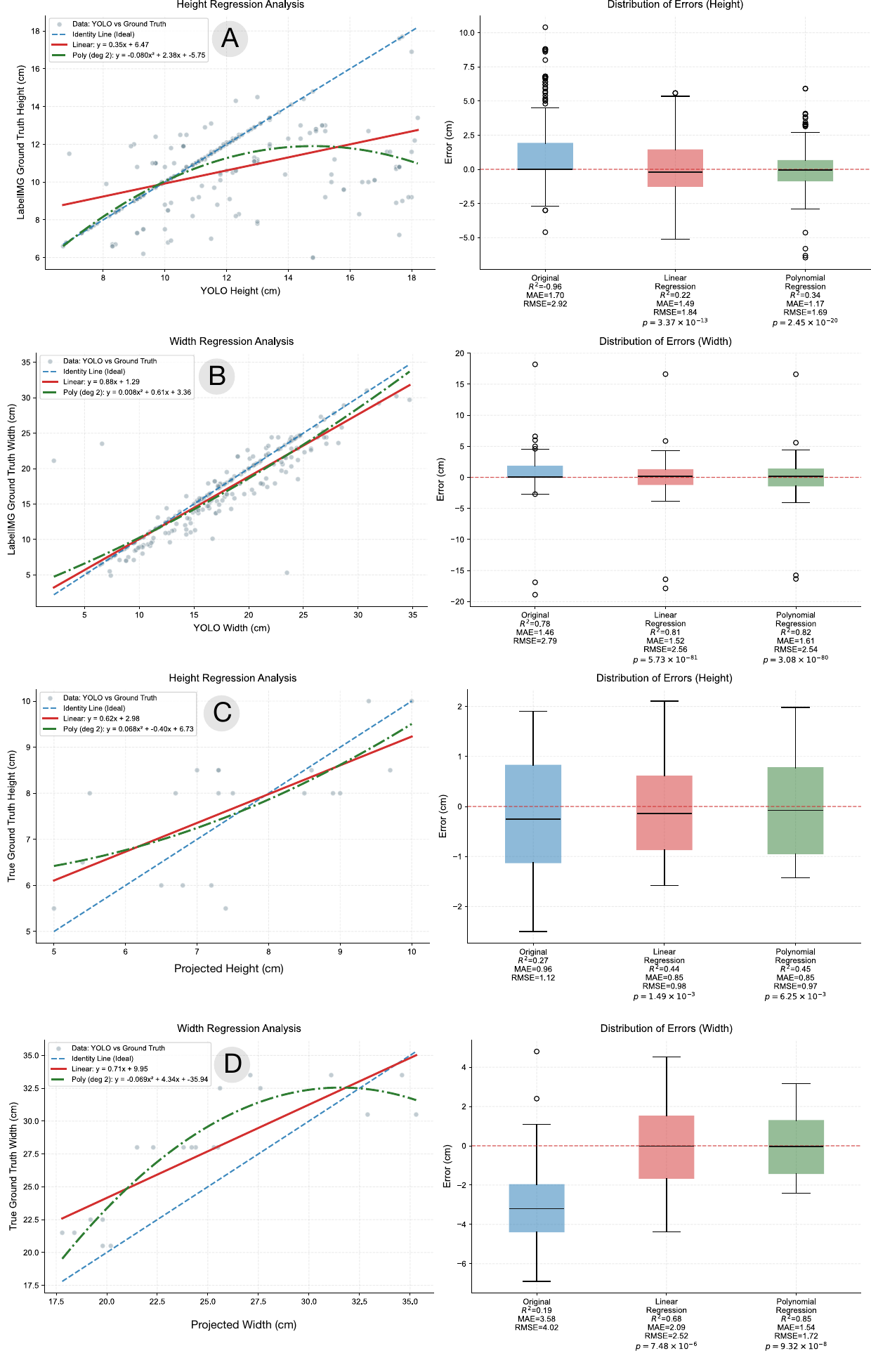}
    \caption{Regression-based correction: (A) Correction of YOLO bounding box shape (height) using simple linear and polynomial regression; (B) Correction of YOLO bounding box shape (width) using simple linear and polynomial regression; (C) Correction of projected bounding box height dimension, by simple linear and polynomial regression; (D) Correction of projected bounding box width dimension, by simple linear and polynomial regression }
    \label{fig12:regression_correction}
\end{figure}

\subsubsection{Bounding box dimensions correction}
\label{sec:size_correction}

Without correction, height predictions demonstrate moderate errors (RMSE = 1.12 cm; MAE = 0.96 cm; $R^2 = 0.27$), with residuals distributed around a slightly negative median of --0.250 cm (Q1 = --1.125 cm; Q3 = 0.825 cm), indicating a marginal tendency to overestimate ground truth heights (Fig.\ref{fig12:regression_correction}c).
Both correction methods yield comparable improvements: linear regression ($y = 0.62x + 2.98$; $R^2 = 0.44$; $p = 1.49 \times 10^{-3}$) reduces RMSE to 0.98 cm and MAE to 0.85 cm, while polynomial regression ($y = 0.068x^2 - 0.40x + 6.73$; $R^2 = 0.45$; $p = 6.25 \times 10^{-3}$) achieves a marginally lower RMSE of 0.97 cm with the same MAE of 0.85 cm.
In both cases, the residuals become slightly more centred: the linear regression yields a median of --0.144 cm (Q1 = --0.862 cm; Q3 = 0.606 cm) and the polynomial regression a median of --0.079 cm (Q1 = --0.944 cm; Q3 = 0.773 cm).
The extremes of the residual distribution are also narrowed, from --2.500 cm / 1.900 cm without correction to --1.581 cm / 2.106 cm under linear regression and --1.422 cm / 1.978 cm under polynomial regression.
Overall, both corrections offer only marginal improvements in height prediction accuracy, suggesting that the geometric model already captures height reasonably well without post-processing.

In contrast, width predictions benefit far more substantially from correction (Fig.~\ref{fig12:regression_correction}d).
Without correction, the model exhibits a strong systematic underestimation of ground truth widths, as evidenced by a median residual of --3.200 cm (Q1 = --4.375 cm; Q3 = --1.975 cm) and an RMSE of 4.02 cm with a MAE of 3.58 cm ($R^2 = 0.19$).
The residual distribution is heavily skewed towards negative values, with the lower whisker reaching --6.900 cm and the upper whisker only 1.100 cm, confirming a near-systematic bias in the uncorrected predictions.
Linear regression ($y = 0.71x + 9.95$; $R^2 = 0.68$; $p = 7.48 \times 10^{-6}$) substantially reduces these errors (RMSE = 2.52 cm; MAE = 2.09 cm) and recentres the residuals (median = --0.013 cm; Q1 = --1.653 cm; Q3 = 1.512 cm), though the interquartile range remains relatively wide and the extremes span --4.371 cm to 4.516 cm.
Polynomial regression ($y = -0.069x^2 + 4.34x - 35.94$; $R^2 = 0.85$; $p = 9.32 \times 10^{-8}$) yields the best results, reducing RMSE to 1.72 cm and MAE to 1.54 cm.
The residuals are effectively recentred (median = --0.030 cm; Q1 = --1.410 cm; Q3 = 1.285 cm) with substantially narrowed extremes (--2.417 cm to 3.168 cm), compared to --6.900 cm / 1.100 cm without 
correction.
This confirms that the relationship between YOLO-predicted and ground truth widths is non-linear, and that polynomial regression is necessary to fully correct the systematic underestimation bias present in the geometric model's width predictions.

\subsubsection{Object size prediction with corrections applied}

For this final step, we adopt an end-to-end evaluation of the complete pipeline applied to the YOLOv11-m model.
Bounding boxes identified by the model are first subject to a shape correction (Section~\ref{sec:bb_correction}), after which the geometric model estimates the corresponding object sizes, which are further refined using the method described in Section~\ref{sec:size_correction}.
Table~\ref{Tab4:yolov11m_train10_pipeline} summarises the mean absolute error (MAE) and root mean square error (RMSE) obtained for each correction strategy, providing a quantitative comparison of their respective impacts 
on estimation accuracy.

For width predictions (Tab.\ref{Tab4a:leak_free_width}), the absence of any correction yields an RMSE of 3.16 cm and a MAE of 2.65 cm.
Applying dimension correction alone already brings substantial improvements: linear dimension correction achieves the best individual correction performance, with an RMSE of 2.23 cm and a MAE of 1.71 cm (highlighted in blue), while polynomial dimension correction yields slightly higher errors (RMSE = 2.33 cm; MAE = 2.08 cm).
Box shape correction alone, whether linear (RMSE = 4.36 cm; MAE = 3.96 cm) or polynomial (RMSE = 4.16 cm; MAE = 3.82 cm), degrades performance relative to the uncorrected baseline, suggesting that shape correction in isolation introduces additional noise into width estimation.
For combined pipeline corrections, the best results are obtained using linear box shape correction followed by polynomial dimension correction, achieving the lowest pipeline RMSE of 2.03 cm and MAE of 1.44 cm 
(highlighted in red).
Polynomial box shape correction combined with linear dimension correction yields a comparable RMSE of 2.27 cm and MAE of 1.93 cm.

For height predictions (Tab.\ref{Tab4b:leak_free_height}), the uncorrected baseline yields an RMSE of 3.57 cm and a MAE of 2.27 cm.
In contrast to width, box shape correction alone proves beneficial for height: polynomial box shape correction achieves the best individual correction performance with an RMSE of 1.79 cm and a MAE of 1.59 cm (highlighted in blue), outperforming linear box shape correction (RMSE = 2.43 cm; MAE = 2.10 cm).
Dimension correction alone also improves results, with linear dimension correction reducing RMSE to 2.18 cm and MAE to 1.62 cm, though polynomial dimension correction alone produces unexpectedly high errors (RMSE = 5.52 cm; MAE = 2.68 cm), suggesting instability when applied without prior shape correction.
For combined pipeline corrections, the optimal strategy applies polynomial box shape correction followed by linear dimension correction, yielding the lowest pipeline RMSE of 1.45 cm and MAE of 1.31 cm (highlighted in red).

The reported RMSE and MAE values should be interpreted within the 
scope of the experimental setup. The evaluation was performed on 20 objects, within a restricted range of sizes and distances (from 1 to 14 meters) from the camera, under relatively stable environmental conditions.
Therefore, these metrics reflect the performance of the method under controlled conditions and may not fully represent its accuracy across a wider range of object geometries and hydrological scenarios.

Overall, dimension-level correction alone is more effective than box shape correction alone for width estimation, while the opposite holds for height estimation.
In both cases, however, the double-level correction pipeline consistently achieves lower errors than any single-level correction strategy.
For width prediction, the best pipeline combines linear box shape correction with polynomial dimension correction (RMSE = 2.03 cm; MAE = 1.44 cm).
For height prediction, the most effective pipeline combines polynomial box shape correction with linear dimension correction (RMSE = 1.45 cm; MAE = 1.31 cm).

\begin{table}[H]
\centering

\begin{subtable}[t]{\textwidth}
\centering
\caption{Width (cm)}
\label{Tab4a:leak_free_width}
\begin{tabular}{lcccccc}
\toprule
 & \multicolumn{2}{c}{No Dim Corr} & \multicolumn{2}{c}{Dim Corr (Lin)} & \multicolumn{2}{c}{Dim Corr (Poly)} \\
\cmidrule(lr){2-3} \cmidrule(lr){4-5} \cmidrule(lr){6-7}
 & RMSE & MAE & RMSE & MAE & RMSE & MAE \\
\midrule
No Box Shape Corr&      3.16&      2.65&      \textcolor{blue}{\textbf{2.23}}&      \textcolor{blue}{\textbf{1.71}}&      2.33&      2.08\\
Box Shape Corr (Lin)&      4.36&      3.96&      2.34&      2.05&      \textcolor{red}{\textbf{2.03}}&      \textcolor{red}{\textbf{1.44}}\\
Box Shape Corr (Poly)&      4.16&      3.82&      2.27&      1.93&      2.1&      1.52\\
\bottomrule
\end{tabular}
\end{subtable}

\vspace{1.5em}

\begin{subtable}[t]{\textwidth}
\centering
\caption{Height (cm)}
\label{Tab4b:leak_free_height}
\begin{tabular}{lcccccc}
\toprule
 & \multicolumn{2}{c}{No Dim Corr} & \multicolumn{2}{c}{Dim Corr (Lin)} & \multicolumn{2}{c}{Dim Corr (Poly)} \\
\cmidrule(lr){2-3} \cmidrule(lr){4-5} \cmidrule(lr){6-7}
 & RMSE & MAE & RMSE & MAE & RMSE & MAE \\
\midrule
No Box Shape Corr&      3.57&      2.27&      2.18&      1.62&      5.52&      2.68\\
Box Shape Corr (Lin)&      2.43&      2.10&      1.76&      1.52&      2.20&      1.86\\
Box Shape Corr (Poly)&      \textcolor{blue}{\textbf{1.79}}&      \textcolor{blue}{\textbf{1.59}}&      \textcolor{red}{\textbf{1.45}}&      \textcolor{red}{\textbf{1.31}}&      1.58&      1.40\\
\bottomrule
\end{tabular}
\end{subtable}

\caption{RMSE and MAE (in cm) under different correction strategies separately for width and height predictions for YOLOv11-m (Train 10). Blue metrics are the lowest RMSE and MAE for individual corrections and red metrics are the lowest RMSE and MAE for pipeline corrections.}
\label{Tab4:yolov11m_train10_pipeline}
\end{table}

To contextualize these results, we introduce the \emph{per-pixel sensitivity} metric, which quantifies the effect of a one-pixel shift in the bounding box on the estimated object size.

\begin{equation}
    S_{\text{pixel}}^{(d)} = \frac{1}{N} \sum_{i=1}^{N} \left| \hat{s}_i - \hat{s}_i^{(d)} \right|
\end{equation}

Here, \( \hat{s}_i \) is the estimated size (width or height) of the \( i \)-th object using the original bounding box, and \( \hat{s}_i^{(d)} \) is the estimated size when the bounding box is shifted by one pixel in direction \( d \), where \( d \in \{ \text{left}, \text{right}, \text{top}, \text{bottom} \} \). The metric \( S_{\text{pixel}}^{(d)} \) represents the average absolute change in predicted size over all \( N \) samples due to this one-pixel perturbation.

This provides a resolution-based lower bound on the estimation error: achieving a lower error than \( S_{\text{pixel}}^{(d)} \) would imply subpixel precision, which is beyond what is expected from a typical image processing algorithm.

By computing \( S_{\text{pixel}}^{(d)} \) for the camera's horizontal and vertical axes on the same dataset, we obtain an average per-pixel sensitivity of \textbf{1.34~cm} for the object width and \textbf{1.36~cm} for the object height.

These values serve as a natural benchmark to interpret the observed RMSE and MAE. Since the reported errors are not substantially greater than these sensitivity thresholds, it indicates that the proposed method leverages nearly all the spatial information available in the input. In practice, this means that the estimated object sizes are nearly as precise as what is physically achievable given the image resolution, leaving little room for further improvement without higher-resolution input data.

\section{Discussion}

\subsection{Inflluence of dataset bias, integrity, and composition on deep learning models}

\subsubsection{Unbalanced dataset}
\label{section:unbalanced_dataset}

A common issue in CNN training is class imbalance, which occurs when one class dominates the dataset while the others are significantly underrepresented \citep{ghosh2024class}.
This imbalance causes the model to prioritise the detection of the dominant class during training \citep{pulgar2017impact}, ultimately leading to a decline in overall performance \citep{masko2015impact}.
Despite containing 6,013 images, our dataset is highly imbalanced, with 10,873 instances of anthropogenic debris compared to just 755 instances of natural debris and 1,010 instances of non-debris materials.
This imbalance, combined with limitations in data quality and setup, reduces the model’s reliability \citep{gong2023survey}.
It also severely affects the model’s ability to generalise, particularly with regard to underrepresented classes \citep{johnson2019survey}.
This highlights the need for larger, more balanced and diverse datasets in order to develop robust and reliable detection models \citep{munappy2022data}.

Beyond class imbalance, the controlled acquisition protocol may also have produced a dataset that is visually cleaner than naturally occurring riverine debris.
In particular, the positive samples were collected during a single experimental campaign, during which debris was manually introduced into the river. 
Whilst this approach ensures controlled conditions and facilitates annotation, it may result in a simplified representation of real-world debris dynamics. 
Indeed, some of the introduced anthropogenic objects may appear less degraded, less fragmented, or more visually contrasted than debris effectively transported by rivers under real environmental conditions. 
More generally, the dataset may underrepresent the variability in object types, degradation states, occlusions, and environmental conditions typically observed in natural river systems, which may affect the generalisation capacity of the trained models. 
At the same time, care was taken to capture the study site under contrasted illumination and sunshine conditions, so as to reproduce part of the environmental variability encountered in practice. 
The resulting dataset should therefore be interpreted as a controlled field benchmark that captures realistic background and lighting variability, but only part of the full visual diversity of riverine debris.
More generally, the dataset should be interpreted as a first benchmark acquired under controlled field conditions, rather than as an exhaustive representation of riverine debris variability.

In this study, this shortcoming is due in particular to in situ data acquisition slowed down by time-consuming manual annotation work (on LabelImg).
A number of approaches can overcome this problem. 
Self-supervised models (SSL) are increasingly used to build large datasets \citep{jaiswal2020survey}. 
Since it does not require annotated data, the SSL model can learn the major characteristics of the input data.
Once the model is pre-trained, it can be reused in a downstream task - such as classification or detection - via a fine-tuning process on a smaller annotated dataset \citep{gui2024survey}. Recently, the use of embeddings from giga-models has emerged as a major breakthrough in deep learning, as it allows models to converge both faster and more effectively. Over the past few months, several notable developments have been introduced, such as AlphaEarth \citep{brown2025alphaearth} for remote sensing data and DINOv3 \citep{simeoni2025dinov3} for photographs, which can enhance performance and reduce false positives for our task.

\subsubsection{Contribution of negative images and effect on the number of false positives}

Object detection in river environments is particularly challenging due to the complexity of captured scenes.
Images taken from fixed cameras on riverbanks, onboard USVs (unmanned surface vehicles; \citealp{bovcon2021mods}) or boats \citep{armitage2022detection} are often affected by water reflectivity, producing glare or mirror-like effects depending on lighting \citep{yalccin2024impact}.
These biases frequently result in false positives, as shown in confusion matrices (Fig.\ref{fig9:yolov8n_results_new} and \ref{fig11:yolov11m_results_new}).
To mitigate this issue, we hypothesised that including negative images (background only) could reduce false positive rates. 
Negative samples were selected under varied environmental conditions: one cloudy day and one bright, clear-sky day at different times of day.
Considering the complex variations that are established in the filmed environment then enables the model to extract more varied background features to account for changing background conditions \citep{gao2019incorporating}.

To reduce false positive rates due to confusion between anthropogenic and natural debris and the river environment, studies are exploring architectural modifications to the models \citep{chang2024modified, tao2024enhanced, wang2024improved} while our work focuses on improving the datasets.

It should be noted that direct comparison with values reported in the literature is complicated by the fact that most published models rely on randomly split datasets, which do not account for spatial or temporal continuity. 
This exposes them to data leakage similar to what we observed in our own baseline experiments, which can substantially affects performance metrics.
The performance of our YOLOv8-n and YOLOv11-m models falls in the range reported in the literature for similar tasks.
Compared to YOLOv8-MSS \citep{wang2024improved}, a model optimised for small object detection through additional dedicated heads (mAP@50 = 87.9\% ; mAP@50-95 = 47.6\%), YOLOv8-n with negative images (Fig.\ref{fig9:yolov8n_results_new} - mAP@50 = 61.8\%; mAP@50-95 = 31.5\%) and YOLOv11-m with negative images (Fig.\ref{fig11:yolov11m_results_new} - mAP@50 = 64.8\%; mAP@50-95 = 40.2\%) achieve lower results.
Other models, such as DSFLDNet, for Dense Small Floating Litter Detection Network, (mAP@50 = 97.3\% ; recall = 93.9\%), which is based on a multidimensional attention mechanism that combines the spatial and frequency domains of the image \citep{ye2025real}, show a superior level of performance to that of YOLOv8-n with negative images (recall = 58.2\%) and YOLOv11-m with negative images (recall = 63\%).

In our experimental setup, both architectures benefit substantially and comparably from the inclusion of 20\% negative images in the leak-free setting.
YOLOv8-n shows a gain of +0.130 in mAP@50 (from 0.488 to 0.618) and +0.153 in precision (from 0.551 to 0.704), while YOLOv11-m achieves comparable improvements of +0.112 in mAP@50 (from 0.536 to 0.648) and +0.213 in precision (from 0.491 to 0.704)(Fig.\ref{fig9:yolov8n_results_new} and \ref{fig11:yolov11m_results_new}).
Both models leverage the additional background diversity provided by the negative images to better constrain their detection boundaries in a complex riverine environment characterised by strong water reflectivity and variable illumination \citep{yalccin2024impact}.
Beyond the 20\% threshold, however, performance declined steadily for both architectures in terms of recall and mAP@50 (Fig.\ref{fig10:ratio_image_neg}).
This can be attributed to a dilution effect: as the proportion of background-only images increases, the relative frequency of annotated object instances in the dataset decreases, reducing the exposure of the model to positive examples during training.
As a result, the model progressively shifts towards a more conservative detection behaviour, manifesting as a systematic drop in recall without a commensurate gain in precision beyond what is already achieved at 20\%.
This trade-off suggests that there exists an optimal operating point specific to the dataset and acquisition conditions, beyond which additional negative images yield diminishing or adverse returns.

However, the acquisition protocol did not include rainfall events. 
This is an important limitation, since intense rainfall often coincides with increased debris inputs in urban rivers and may also alter image quality through reduced visibility, raindrop artefacts, stronger surface turbulence, and rapidly changing illumination conditions. 
Extending the dataset to rainy-weather situations therefore appears to be an important next step for improving the operational robustness of the proposed framework.

\subsubsection{Effect of training on the leak-free dataset}

In machine learning (ML), data leakage occurs when some of the data in the training set also appears in the validation or test sets \citep{tampu2022inflation, john2025problematic}.
While this does not directly affect the internal learning process of the model, it artificially inflates performance metrics, providing an inaccurate impression of the model's effectiveness in real-world scenarios \citep{apicella2024don}.
Leakage can arise from pre-processing steps or from the inherent structure of the data itself \citep{seide2011feature}.
In the present study, image data are extracted from continuous video footage capturing the periodic occurrence of anthropogenic and natural debris as well as non-debris material.
Similar images from the same temporal scene can be found in both training and test sets when the dataset is split into three parts (training, validation and test; \citep{bouke2023empirical}), as was the case from Train 1 to 6 (Tab.\ref{tab1:training_model}).

As the data is acquired in the form of video time series, the objective of distinguishing images from different recording scenes implies their transformation into embedded vectors. 
These vectors combine visual features extracted using a YOLOv8-n encoder, bounding box information and the corresponding timestamps.
This representation reduces the dimensionality of the data while retaining its semantic richness \citep{figueiredo2024analyzing}. 
A t-SNE is generated from the embedded vector (used to reduce the data's dimensionality) in order to facilitate visualisation and apply DBSCAN clustering (Fig.\ref{fig8:t-SNE}). 
While t-SNE is particularly well suited to data with a temporal component, other methods, such as principal component analysis (PCA), can be used for preprocessing large datasets acquired under varying conditions (i.e. scenes, objects and sensors \citep{sasse2025overview}).
This method eliminates the need for additional information such as timestamps. 
Clustering methods can be applied to the PCA results to identify relevant groupings once the data has been projected into a reduced-dimensional space.
These groupings can then be used to create separate data sets (i.e. training, validation, testing) and avoid time leaks or interset redundancies.
This type of strategy is particularly useful in cases of heterogeneous acquisition (different sensors, lighting conditions, camera angles), in preparation for cross-validation procedures or successive implementations \citep{aarnink2025automatic}.

The CNNs that were trained using the dataset without any data leakage, namely YOLOv8-n (precision = 55.1\%; mAP@50 = 48.8\%) and YOLOv11-m (precision = 49.1\%; mAP@50 = 53.6\%), demonstrate lower detection performance than some of the models that have been reported in the literature.
For instance, YOLOv5 models designed specifically for wood detection in rivers achieve much higher precision (99.5\%; \citep{donal2023automated}).
Similarly, YOLOv5 models designed for detecting anthropogenic debris report mAP@50 scores of 88\% \citep{nunkhaw2024image} and an precision of 86\% \citep{luo2022water}.
However, while the models developed in this study may underperform in comparison, their metrics are more realistic as they are not inflated by data leakage.
In contrast, models from the literature may produce overestimated results due to random dataset splits that do not account for spatial or temporal continuity, leading to similar images appearing in both the training and test sets.
Furthermore, the robustness and generalisability of detection models depend on data acquisition conditions, such as environmental variability and sensor placement. 
This further supports the need for cautious interpretation of high performance metrics when data leakage risks are not explicitly addressed.

This approach is a replicable methodology for reducing data leakage when working with time-series data.
In the context of a restricted dataset with little variation in recording conditions, complementary information such as bounding boxes and timestamps is necessary.
For these data, we recommend using an embedded vector followed by t-SNE to differentiate between scenes and divide them into groups, creating several datasets.
In case of larger and more heterogeneous datasets, it is preferable to apply PCA followed by a clustering method based solely on the visual characteristics of the images to minimise the presence of similar images in different datasets.

\subsubsection{Effect of class weighting on detection performance}

Class weighting was investigated as a complementary strategy to further improve the detection of underrepresented categories, particularly class 2 (natural debris), which exhibited persistently high false negative rates across all leak-free configurations \citep{johnson2019survey}.

For both architectures, class weighting consistently improved the true positive rate for class 2: from 0.45 to 0.59 for YOLOv8-n, and from 0.48 to 0.83 for YOLOv11-m (Fig.~\ref{fig9:yolov8n_results_new} and \ref{fig11:yolov11m_results_new}).
These results indicate that increasing the contribution of minority classes in the loss function effectively enhances their detectability.
However, these gains come at the cost of a notable increase in false positives in background regions, as well as a reduction in global performance metrics \citep{phan2020resolving}.
Specifically, mAP@50 decreases from 0.618 to 0.594 for YOLOv8-n and from 0.648 to 0.614 for YOLOv11-m. This behaviour reflects a fundamental precision–recall trade-off.

From an optimisation perspective, class weighting modifies the training objective by increasing the penalty associated with the misclassification of underrepresented classes \citep{crasto2024class, ghosh2024class}. 
Consequently, the model becomes more sensitive to these classes and produces more detections, even in uncertain or ambiguous regions. While this reduces false negatives, it also increases false positives, favouring less confident detections.
This effect is particularly evident in the confusion matrices, where background regions are more frequently misclassified as objects, especially for class 2.
This indicates that the model is less restrictive and more prone to detecting weak or ambiguous visual patterns.

From a practical perspective, the 20\% negative images configuration without class weighting remains the preferred strategy when global detection accuracy is the primary objective. Class weighting may, however, be preferable in monitoring scenarios where minimizing missed detections of natural debris is critical, for instance when estimating total debris flux or distinguishing debris types for downstream mass estimation.

Overall, these results highlight that class weighting does not uniformly improve detection performance, but rather shifts the balance between recall and precision, making it a task-dependent optimization choice.

\subsection{Object size prediction}

In our study area, the absence of man-made structures (such as bridges) justifies the installation of cameras on the riverbank.
This installation leads to geometric deformation (perspective) due to the camera's inclined position, which is not the case when a camera is fixed to a bridge (providing a vertical view; \citep{van2020automated}).
To correctly estimate the dimensions of debris, it is necessary to correct geometric distortion, thereby limiting errors in predicted heights and widths.

The proposed pipeline achieves a high level of precision in estimating real-world object dimensions, despite numerous sources of visual and geometric uncertainty in outdoor river imagery.
Through the combination of a physics-based geometric correction and a statistical calibration step, the method consistently yields dimension estimates that are sufficiently accurate for downstream applications such as plastic mass estimation. 
This section discusses the model's robustness, underpinning assumptions, inherent and practical limitations, and perspectives for future work.

\subsubsection{Geometric model and approximation assumptions}

The model uses camera parameters (i.e. installation height, tilt angle, and focal length) to project bounding box coordinates from image space to physical coordinates. Without requiring any assumption on object orientation or planarity, the model achieves accurate results(Tab.\ref{Tab4:yolov11m_train10_pipeline}). 
For instance, width RMSE improves from 3.16~cm (uncorrected) to 2.03~cm after linear and polynomial correction (Tab.\ref{Tab4a:leak_free_width}), while height estimates are refined to within 1.45~cm (Tab.\ref{Tab4b:leak_free_height}).

The model includes certain approximations that rely on simplifying geometric assumptions, which, while not strictly exact, are operationally effective.
First, it assumes the object center lies along the ray passing through the center pixel of the bounding box. 
This assumption provides a stable basis for geometric projection in practice, even if it is not theoretically justified.
Second, the object size is estimated by summing two ray-plane intersections (top/bottom and left/right). 
While these are only approximately valid, they yield accurate results due to the near-parallel nature of rays in the camera configuration employed.

It is worth noting that the computed \emph{height} corresponds to the projection of the object along the camera’s vertical viewing axis, not a strict vertical measurement in world coordinates.
In most cases, the distinction is negligible, particularly for compact floating debris, but the difference may become relevant for extended or tilted objects (i.e. elongated wooden logs).
Nevertheless, our empirical corrections compensate for such biases within the current application scope.

\subsubsection{Limitations of bounding box representation and prospects for rotation-aware detection}

The present pipeline operates with axis-aligned bounding boxes, which introduces a practical limitation when estimating the dimensions of rotated objects (Fig.\ref{fig14:bottle_biais}).
A tilted object enclosed in an axis-aligned bounding box results in a distorted representation: the box tends to underestimate the object’s major dimension and overestimate its minor one, effectively squaring the projection.
However, this is a constraint of the detector representation rather than the geometric model itself. 
The projection algorithm is capable of handling rotated boxes and would presumably produce correct dimension estimations given accurate angular information.

\begin{figure}[H]
    \centering
    \includegraphics[width=1\linewidth]{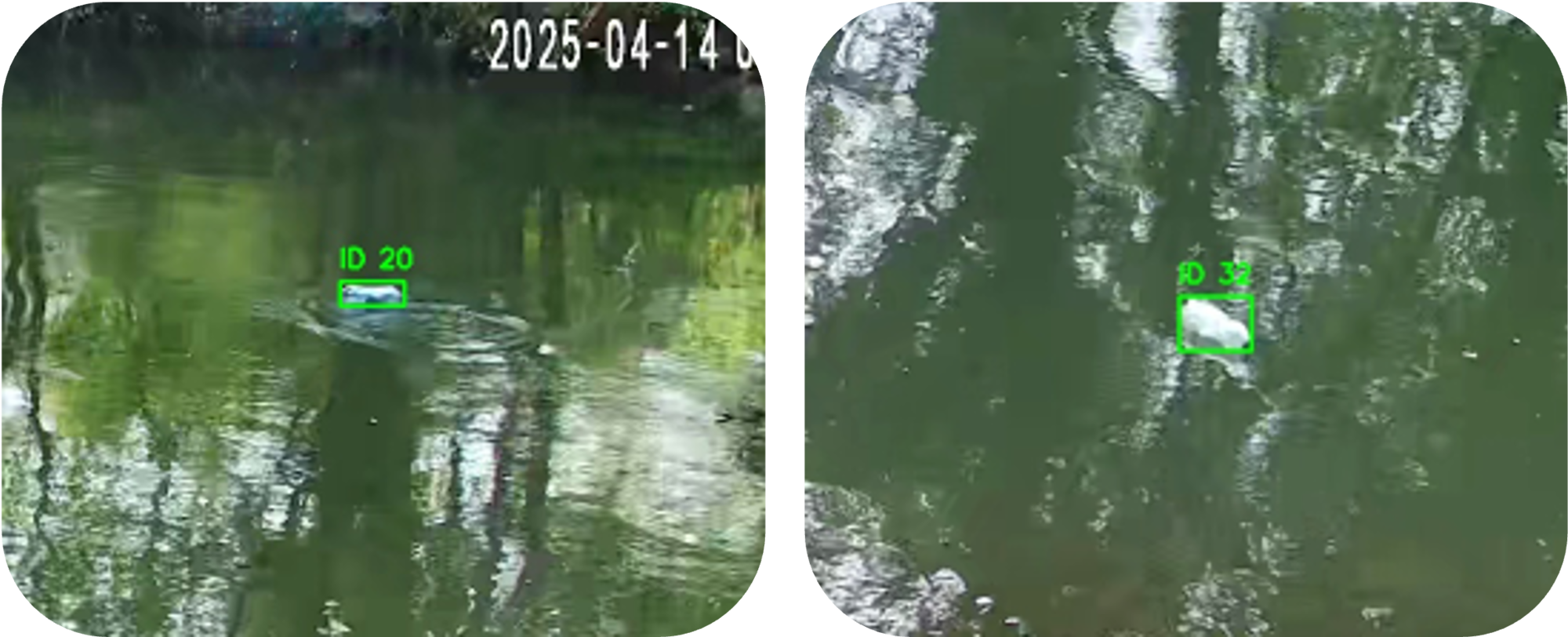}
    \caption{Example of parallel (left) and inclined (right) bottle}
    \label{fig14:bottle_biais}
\end{figure}

Recent object detectors support rotated bounding boxes (i.e., YOLOv11-OBB; \citep{ultralytics2024obb}), and future integration of such models is expected to entirely mitigate this limitation. 
Accordingly, object rotation should not be considered as a fundamental barrier but rather as a temporary constraint inherent to the current detection model. 
Our evaluation has therefore focused on objects oriented either horizontally or vertically to avoid bias from this artifact.

\subsubsection{Error analysis and proximity to resolution limits}

Following correction, the residual estimation errors approach the sensitivity threshold dictated by the image resolution and scene geometry.
We define a per-pixel sensitivity metric—roughly 1.3 cm for both width and height under our conditions—as a lower bound on attainable precision. 
The final RMSEs of 2.03 cm (width) and 1.45~cm (height) for the leak-free model approach this limit, indicating that the method extracts nearly all the geometric information available from the image.

This suggests that further accuracy improvements would likely require higher-resolution sensors, more precise calibration, or alternative imaging modalities.
Given that our objective is to provide a lightweight and field-deployable system, the current error range is both satisfactory and near-optimal in the given constraints.

A limitation of the size-estimation experiment lies in the restricted diversity of the evaluation dataset. 
The objects considered cover a limited range of shapes, sizes, and distances, and were acquired under relatively stable water-level conditions.

In real-world river environments, object morphology, orientation, occlusions, and hydrological conditions may vary significantly, which may significantly affect estimation accuracy.
Further experiments would be required to assess the robustness of the method across a broader range of conditions.

A secondary constraint pertains to the spatial extent of the evaluation. 
All quantitative size-estimation experiments were conducted on a single river and camera configuration (Steingiessen).

Despite the fact that the geometric model is parameterised using camera intrinsic and extrinsic parameters, which renders it theoretically transferable, no quantitative validation was performed on independent sites or acquisition setups.

Consequently, the robustness of the size-estimation pipeline under varying environmental conditions (i.e., different river widths, camera angles, water levels, or object distributions) remains to be assessed.

\subsubsection{Environmental artifacts: reflections and occlusions}

Natural environments introduce unavoidable artifacts, including water-surface reflections and occlusions. 
Debris reflections on water surface \citep{yalccin2024impact} often extend the vertical footprint of the object in the image, leading to overestimated height predictions. 
Occlusions may result in truncated boxes, contributing to underestimation (Fig.\ref{fig15:dimension_correction_water} b).

These phenomena particularly affect the Y-axis of bounding boxes and can explain some systematic biases in height estimation. 
While our corrections alleviate these issues to some extent, they cannot fully remove the underlying visual ambiguity.
Additional methods such as temporal filtering or reflection-aware postprocessing could further improve accuracy in these scenarios.

\begin{figure}[H]
    \centering
    \begin{subfigure}[b]{1\textwidth}
        \centering
        \includegraphics[width=\textwidth]{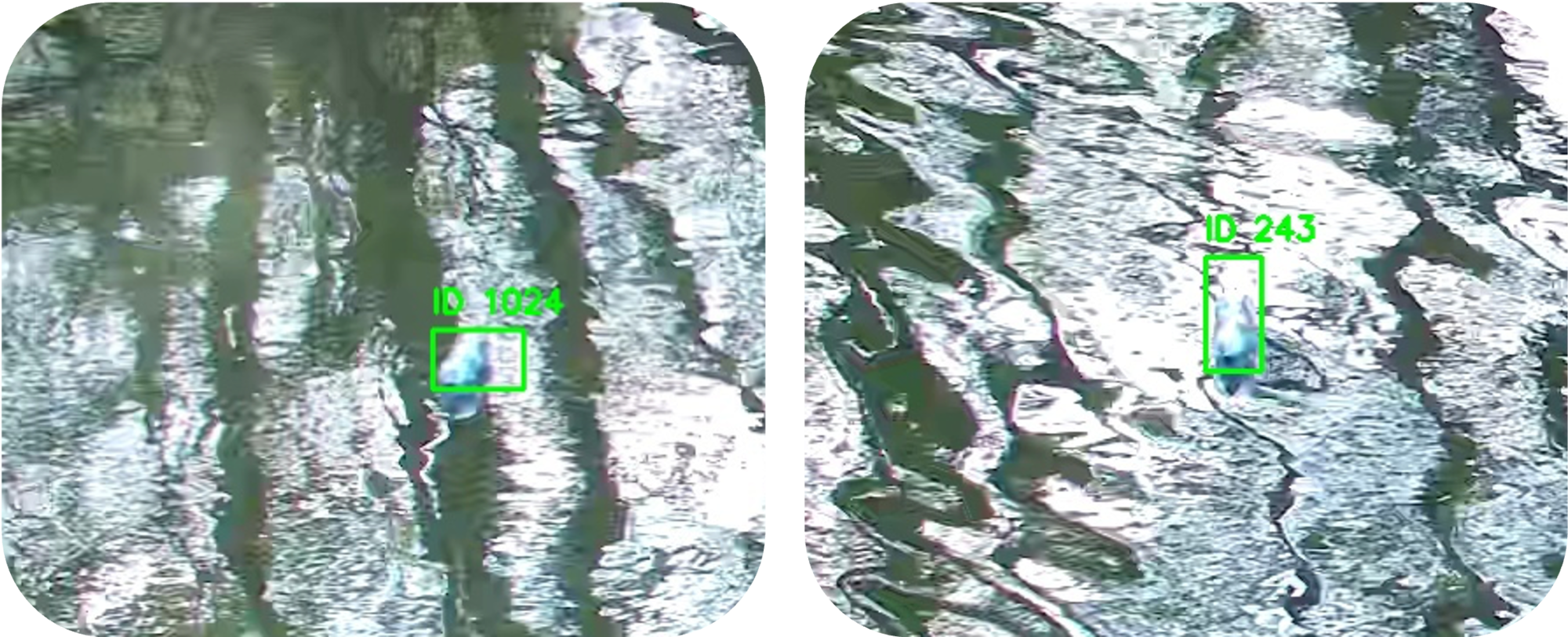}
        \caption{Light reflection on the surface of the water}
        \label{fig15a:mirror_effet}
    \end{subfigure}
    
    \vspace{0.5cm} 

    \begin{subfigure}[b]{1\textwidth}
        \centering
        \includegraphics[width=\textwidth]{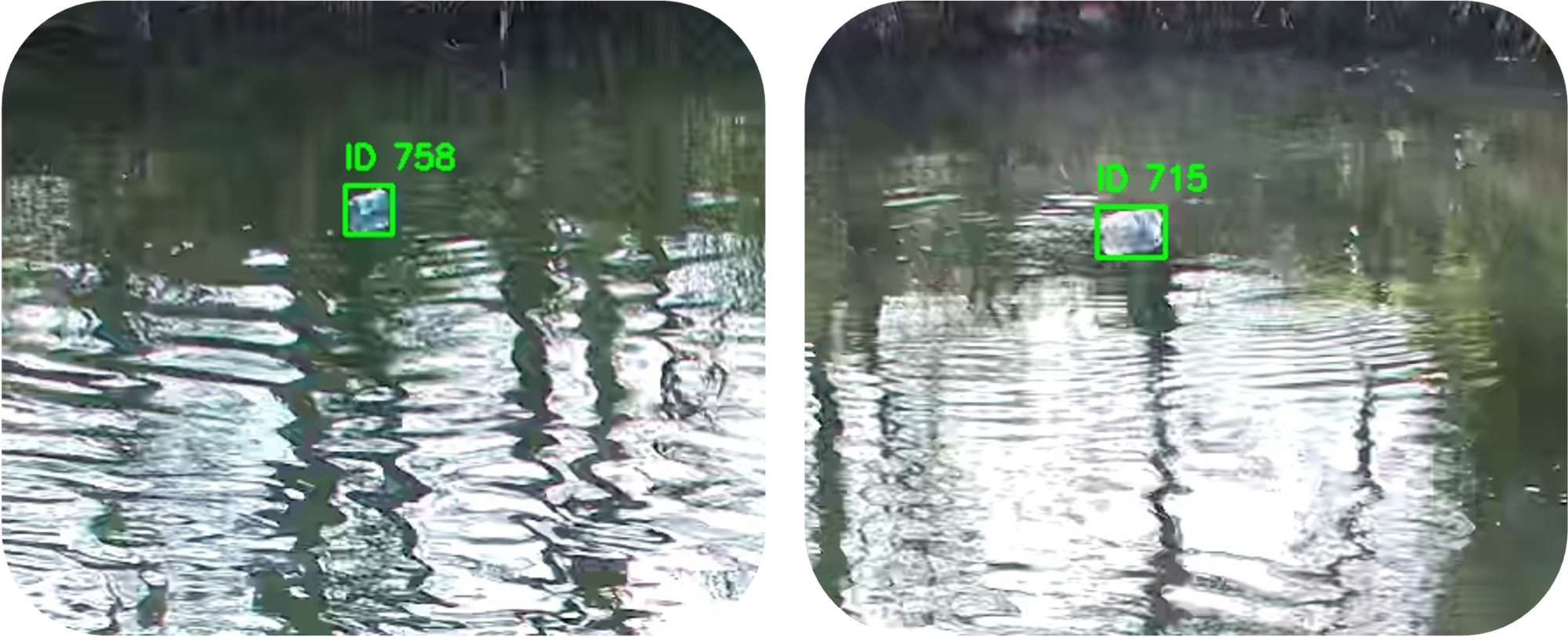}
        \caption{Bottles reflected on the water}
        \label{fig15b:bottle_reflect}
    \end{subfigure}
    
    \caption{Examples of mirror effect of water}
    \label{fig15:dimension_correction_water}
\end{figure}

\subsubsection{Morphological asymmetries and object orientation}

Shape-induced asymmetries presumably contribute to residual width estimation error.
For instance, rotating a horizontally placed bottle by 180° can yield a different bounding box due to the tapering near the cap. 
This asymmetry, combined with axis-aligned detection, introduces a measurable variation in estimated size.
Future detectors that support instance-level segmentation or rotated bounding boxes may help reduce this variability.

\subsection{First spatial inference testing works of YOLOv11-m to detect plastic debris on the Bruche river}

In order to provide an initial qualitative indication of cross-site detection transferability beyond the main study setup, the YOLOv11-m model was inferred on images obtained from the Bruche River.

The Bruche is an 80-kilometre-long Rhine sub-tributary that originates in the Vosges Mountains, and subsequently drains the Upper Rhine Graben \citep{jautzy2020measuring}. 

A 3-megapixel pan-tilt-zoom (PTZ) camera was installed on the DREAL (Direction Régionale de l'Environnement, de l'Aménagement et du Logement) Holtzheim hydrometric station on the left bank of the river (Fig.\ref{fig16:bruche_river}).
A qualitative analysis of YOLOv11-m's performance on these images highlights the model's ability to detect anthropogenic debris and non-debris materials under various recording conditions. 
These spatial inference results are encouraging as the model was originally trained on a different river (the Steingiessen).

As previously discussed (Section \ref{section:unbalanced_dataset}), increasing both the volume and diversity of training data enhances detection performance.
The deployment of supplementary cameras appears to be a pertinent strategy for the acquisition of such enriched datasets.

It is imperative to acknowledge the limitations of this experiment, which is constrained to a qualitative evaluation of detection transferability.

\begin{figure}[H]
    \centering
    \includegraphics[width=1\linewidth]{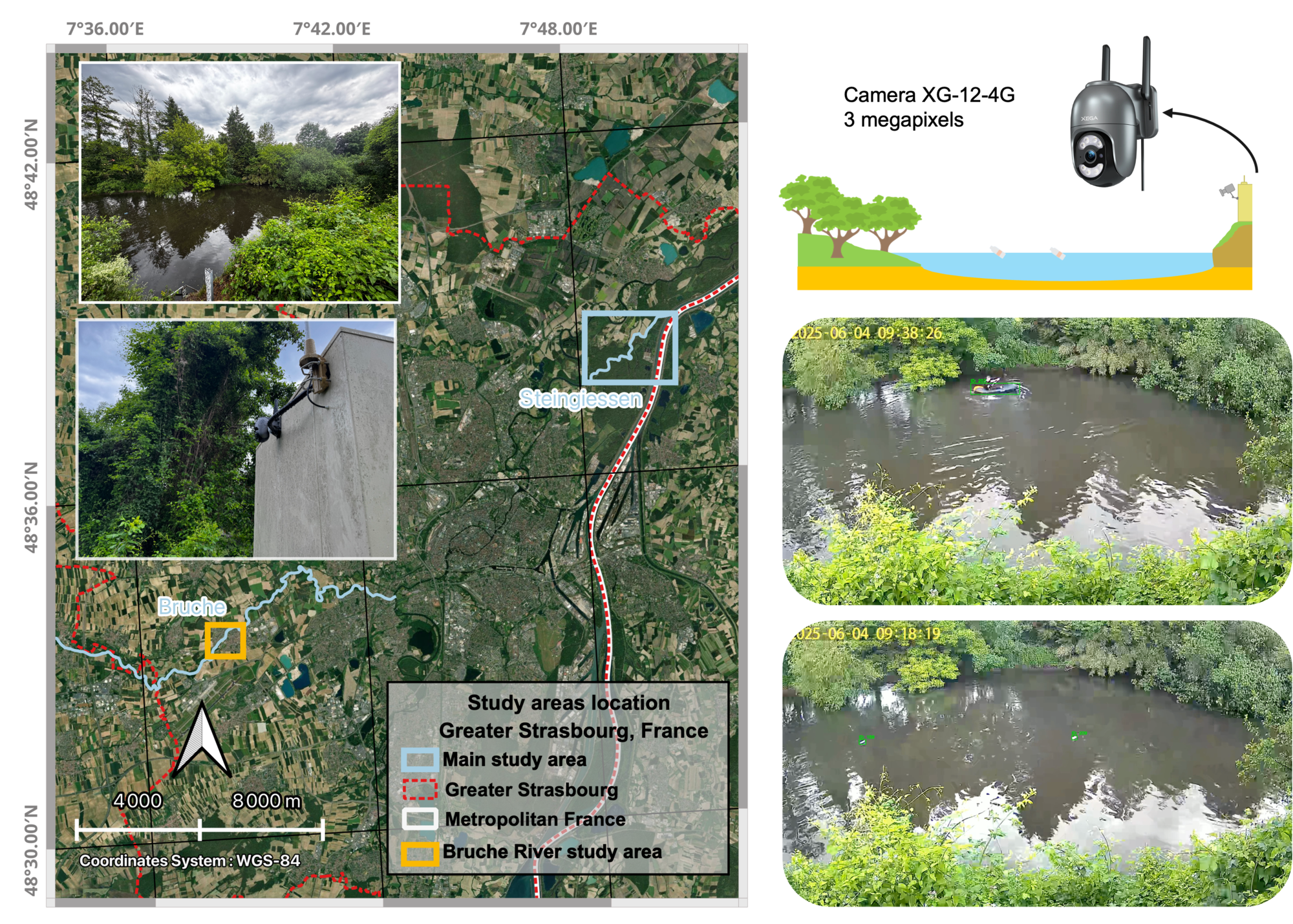}
    \caption{Spatial inference of YOLOv11-m (Train 10) model on Bruche River (South-West Greater Strasbourg)}
    \label{fig16:bruche_river}
\end{figure}

\subsection{Contribution to the Overarching Research Goal}

This study contributes by combining three aspects that are rarely addressed together: dataset biases, including negative images, temporal leakage and class weighting; detection under oblique camera configurations; and an interpretable geometric pipeline for object size estimation.

Each of these has been explored separately, and their combined evaluation provides new insights into deploying camera-based river monitoring systems.

\subsubsection{Models for detecting anthropogenic debris in rivers}

In accordance with the objectives of our experimental benchmark, the YOLOv8-n and YOLOv11-m models demonstrated the most significant trade-offs among the architectures that were examined.
This benchmark should be regarded as a preliminary context-specific comparison of object detection architectures for floating anthropogenic debris under controlled yet realistic field conditions.

The compared analysis of two models with contrasting degree of complexity in terms of architecture is a first step towards operational issues. 
Future studies should foster the creation of models embedded in cameras to create a tool that can be deployed at the catchment scale to quantify the global phenomenon of plastic pollution.

\subsubsection{A tool for estimating the dimensions of objects detected by a monocular camera}

The proposed method demonstrates that reliable, centimeter-level object dimension estimation is feasible with a monocular bank-mounted camera, provided that the imaging parameters are known and a suitable geometric model is applied.
The performance is close to the theoretical limits set by image resolution and projection geometry.

By explicitly modeling perspective, correcting for biases, and acknowledging practical limitations, the pipeline offers an interpretable and effective tool to monitor riverine debris. 

It constitutes a promising foundation upon which future refinements—such as orientation-aware detection, reflection filtering, or mass estimation models—can be built.

However, the quantitative evaluation of this approach remains limited to a single experimental setup, and further validation across multiple sites and camera configurations is required to confirm its general applicability.

\subsubsection{From size to mass estimation: assumptions and limitations}

While the current pipeline estimates object dimensions only, the broader goal is to infer plastic mass or volume.
The implicit assumption is that object size correlates with plastic content, particularly for bottles.
This is a reasonable first approximation; however, it may not hold for atypical objects or classification errors. 
For example, a kayak misclassified as a large plastic item could substantially bias aggregate mass estimates.

In future work, size-based consistency checks will be incorporated to reject implausible combinations of classification and dimension.
More refined estimations would also benefit from integrating material recognition or object-type-specific volume models.

\section{Conclusion}

This study proposes an innovative and reproducible framework for 
monitoring anthropogenic floating waste in rivers, combining deep 
learning-based object detection and geometric size estimation from 
monocular bank-mounted cameras, with three main contributions.

The first contribution is a reproducible experimental framework for 
investigating dataset biases in river debris detection.
A DBSCAN-based temporal clustering strategy was introduced to construct 
a leak-free dataset from video-derived image sequences, revealing that 
data leakage inflates performance metrics by more than 40 percentage 
points in mAP@50 relative to leak-free conditions.
The systematic evaluation of negative image proportions identified 20\% 
as the optimal operating point, beyond which a dilution effect 
degrades recall without further precision gains.
Class weighting was shown to improve the detection of underrepresented 
natural debris at the cost of increased false positives, constituting 
a task-dependent trade-off rather than a universal improvement.
Taken together, these results highlight that dataset construction 
choices — leakage control, negative image proportion, and class 
weighting — have a greater impact on reported performance than 
architectural differences between models.

The second contribution is an evaluation of two YOLO architectures of 
contrasting complexity under bank-mounted, oblique-view conditions.
A comparative evaluation of several YOLO variants revealed that 
YOLOv8-n and YOLOv11-m exhibited effective performance under the 
specific conditions investigated in this study. Instead of establishing 
a universally applicable ranking of detection models for riverine debris 
monitoring, these results underscore context-dependent trade-offs 
between detection accuracy, inference speed, and robustness to dataset 
construction effects.
Under leak-free conditions, YOLOv11-m consistently outperformed 
YOLOv8-n across all configurations, and a preliminary inference test 
on the Bruche River provided encouraging qualitative evidence of 
cross-site transferability.
Future work will focus on extending the dataset across multiple 
sites, seasons, and hydrological conditions — including rainfall events, 
which were absent from the current acquisition protocol — in order to 
better capture the natural variability of riverine debris and improve 
the generalisation capacity of the proposed framework.

The third contribution is an interpretable monocular geometric pipeline 
for estimating object dimensions from detections.
By combining a physics-based perspective projection model with a 
double-level statistical correction, the pipeline achieves final RMSEs 
of 1.45~cm (height) and 2.03~cm (width), approaching the theoretical 
precision limit imposed by image resolution (approximately 1.3~cm per 
pixel), and adequate for downstream mass or volume estimation.
It constitutes a promising foundation upon which future 
refinements — such as orientation-aware detection, reflection 
filtering, or mass estimation models — can be built.
However, the quantitative evaluation of this approach remains 
limited to a single experimental setup, and further validation across 
multiple sites and camera configurations is required to confirm its 
general applicability.

Overall, this work provides a solid methodological foundation for the 
development of robust, low-cost, automated monitoring systems for river 
debris, with a view to operational deployment across entire watersheds.
\section*{CRediT authorship contribution statement}

\textbf{Gauthier Grimmer}: Conceptualization, Methodology, Investigation, Software, Data curation, Writing - Original Draft, Writing - Review \& Editing. \textbf{Romain Wenger}: Conceptualization, Methodology, Software, Formal analysis, Writing - Original Draft, Writing - Review \& Editing, Validation, Funding acquisition, Supervision. \textbf{Clément Flint}: Methodology, Software, Writing - Original Draft, Writing - Review \& Editing, Validation. \textbf{Germain Forestier}: Validation, Formal analysis, Writing - Review \& Editing, Supervision. \textbf{Gilles Rixhon}: Validation, Writing - Review \& Editing, Supervision. \textbf{Valentin Chardon}: Conceptualization, Formal analysis, Methodology, Writing - Original Draft, Writing - Review \& Editing, Supervision, Validation, Funding acquisition, Project administration. 

\section*{Declaration of competing interest}

The authors declare that they have no known competing financial interests or personal relationships that could have appeared to influence the work reported in this paper.

\section*{Data availability}

The dataset is available on demand by contacting the authors. The code is available on Github (Projective geometry: \url{https://github.com/cflinto/projective-geo-sizing} ; deep learning models: \url{https://github.com/g-grimmer/YOLO-WastDetect/} ; corresponding models weights: \url{https://cloud.live.unistra.fr/index.php/s/MLQKkXPDaHWZH7D}).

\section*{Acknowledgements}

This research has been funded by the \textit{Zone Atelier Environnementale et Urbaine} of Strasbourg. \\
We also would like to thank the University of Strasbourg through the \textit{Ecole Doctorale Sciences de la Terre et de l’Environnement (ED n° 413)} for the PhD grant. \\
Thank you to David Eschbach and all the colleagues from the \textit{Eurométropole de Strasbourg} for the support and the funding of the project. \\
We also thank Anne Puissant and Pierre-Alexis Herrault for their feedbacks.
The authors would also like to acknowledge the High Performance Computing Center of the University of Strasbourg for supporting this work by providing scientific support and access to computing resources. Part of the computing resources were funded by the Equipex Equip@Meso project (Programme Investissements d'Avenir) and the CPER Alsacalcul/Big Data. \\
The authors would like to thank the DREAL Grand Est for their invaluable operational support in the field. \\
Finally, we would like to thank the anonymous reviewers and the editorial board for their insightful comments and suggestions, which greatly improved the quality of this manuscript.

\bibliographystyle{elsarticle-harv} 
\bibliography{bib}

\newpage

\section*{Appendix A}\label{appendix:A}

\paragraph{Plane from a point and two non-colinear vectors}

In three-dimensional space, a plane can be defined by a unit normal vector \( \vec{n} \) and a scalar offset \( d \), such that every point \( \vec{x} \) on the plane satisfies the equation:
\begin{align}
\vec{n} \cdot \vec{x} + d = 0.
\end{align}
This is known as the \emph{implicit form} of a plane equation~\citep{do2016differential}.

To construct such a plane from geometric data, consider a known point \( \vec{P}_0 \in \mathbb{R}^3 \) that lies on the plane, along with two non-colinear vectors \( \vec{v}_1 \) and \( \vec{v}_2 \) that span the plane.

We compute the normal vector \( \vec{n} \) as the normalised cross product of these vectors:
\begin{align}
\vec{n} = \text{normalise}(\vec{v}_1 \times \vec{v}_2).
\end{align}
Then, the offset \( d \) is determined so that the plane passes through \( \vec{P}_0 \), yielding:
\begin{align}
d = -\vec{n} \cdot \vec{P}_0.
\end{align}

\paragraph{Intersection of a ray with a plane}

A ray in 3D space is defined by an origin point \( \vec{o} \in \mathbb{R}^3 \) and a direction vector \( \vec{v} \in \mathbb{R}^3 \), and is parameterized as:
\begin{align}
\vec{r}(t) = \vec{o} + t \vec{v}
\quad \text{for } t \geq 0.
\end{align}
The ray extends from \( \vec{o} \) in the direction of \( \vec{v} \), and the parameter \( t \) controls the distance traveled along the ray.

To compute the intersection point \( \vec{p} \) between the ray and the plane, we substitute the ray equation into the plane equation:
\begin{align}
\vec{n} \cdot (\vec{o} + t_{intersection} \vec{v}) + d = 0.
\end{align}
Solving for \( t \) gives:
\begin{align}
t_{intersection} = -\frac{\vec{n} \cdot \vec{o} + d}{\vec{n} \cdot \vec{v}}
\quad \text{provided that } \vec{n} \cdot \vec{v} \ne 0.
\end{align}
If the resulting \( t_{intersection} \geq 0 \), the intersection lies in the forward direction of the ray and is considered valid. The corresponding 3D intersection point is then:
\begin{align}
\vec{p} = \vec{o} + t_{intersection} \vec{v}.
\end{align}

\paragraph{Distance from a point to a plane}

Given a plane \( \vec{n} \cdot \vec{x} + d = 0 \) and a point \( \vec{p} \in \mathbb{R}^3 \), the signed distance from the point to the plane is:

\begin{align}
\text{distance} = \vec{n} \cdot \vec{p} + d.
\end{align}

The geometric (unsigned) distance is given by:
\begin{align}
\text{distance} = \left| \vec{n} \cdot \vec{p} + d \right|,
\end{align}
which corresponds to the shortest distance between the point and any location on the plane.

\section*{Appendix B}\label{appendix:B}

\begin{table}[h]
\centering
\caption{Detailed ground-truth object dimensions used for validation.}
\label{tab:size_detailed}
\begin{tabular}{ccc}
\hline
\textbf{Object ID} & \textbf{Width (cm)} & \textbf{Height (cm)} \\
\hline
1 & 28.0 & 8.0 \\
2 & 28.0 & 8.0 \\
3 & 28.0 & 8.0 \\
4 & 28.0 & 8.0 \\
5 & 28.0 & 8.0 \\
6 & 28.0 & 8.0 \\
7 & 28.0 & 8.0 \\
8 & 32.5 & 10.0 \\
9 & 32.5 & 10.0 \\
10 & 33.5 & 8.5 \\
11 & 20.5 & 6.0 \\
12 & 21.5 & 5.5 \\
13 & 21.5 & 5.5 \\
14 & 22.5 & 6.0 \\
15 & 22.5 & 6.5 \\
16 & 33.5 & 8.5 \\
17 & 20.5 & 6.0 \\
18 & 30.5 & 8.5 \\
19 & 33.5 & 8.5 \\
20 & 30.5 & 8.5 \\
\hline
\end{tabular}
\end{table}






\end{document}

%% file: figure/method/fig_screen_box.tikz.tex
\begin{tikzpicture}[x=1.2cm,y=1.2cm,scale=1,transform shape]

\node[anchor=south west, inner sep=0] (img) at (0,0) {\includegraphics[width=6cm]{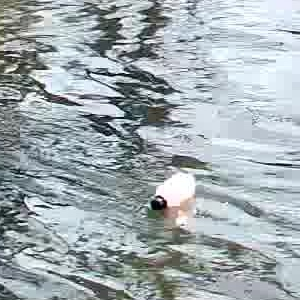}};

\begin{scope}[shift={(0,0)}, x={(img.south east)}, y={(img.north west)}]
    \draw[red, very thick] (0.475,0.25) rectangle (0.675,0.45);

    \draw[->, thick] (-0.1, -0.1) -- (1.1, -0.1) node[right] {$x_{\text{pix}}$};
    \draw[->, thick] (-0.1, -0.1) -- (-0.1, 1.1) node[above] {$y_{\text{pix}}$};

\end{scope}

\end{tikzpicture}

%% file: figure/method/fig_projected_vectors.tikz.tex
\begin{tikzpicture}[scale=1.8, every node/.style={font=\scriptsize}]
    \coordinate (C) at (0.6,1.0);

    \coordinate (TL) at (1.9,1.1);
    \coordinate (TR) at (1.225,1.0);
    \coordinate (BR) at (1.15,0.3);
    \coordinate (BL) at (1.825,0.4);

    \coordinate (TLBB) at (1.725, 0.75);
    \coordinate (TRBB) at (1.5, 0.725);
    \coordinate (BRBB) at (1.475, 0.525);
    \coordinate (BLBB) at (1.7, 0.55);

    \draw (TL) -- (TR) -- (BR) -- (BL) -- cycle; 
    \draw[-, black] (C) -- (TL);
    \draw[-, black] (C) -- (TR);
    \draw[-, black] (C) -- (BR);
    \draw[-, black] (C) -- (BL);

    \node[anchor=center, inner sep=0] (img) at (1.525,0.7) {\includegraphics[width=1.3cm]{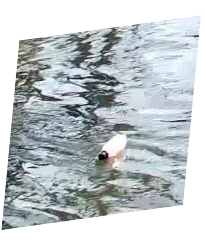}};

    \draw[red, thick] (TLBB) -- (TRBB) -- (BRBB) -- (BLBB) -- cycle; 

    \draw[red, line width=0.4pt, dash pattern=on 2pt off 1.5pt] (C) -- (TLBB);
    \draw[red, line width=0.4pt, dash pattern=on 2pt off 1.5pt] (C) -- (TRBB);
    \draw[red, line width=0.4pt, dash pattern=on 2pt off 1.5pt] (C) -- (BRBB);
    \draw[red, line width=0.4pt, dash pattern=on 2pt off 1.5pt] (C) -- (BLBB);

    \draw[->, red] (TLBB) -- ($(C)!1.7!(TLBB)$);
    \draw[->, red] (TRBB) -- ($(C)!1.7!(TRBB)$);
    \draw[->, red] (BRBB) -- ($(C)!1.7!(BRBB)$);
    \draw[->, red] (BLBB) -- ($(C)!1.7!(BLBB)$);

    \node[red, font=\scriptsize] at (2.25,0.8) {$\vec{v}_{\text{top-left}}$};
    \node[red, font=\scriptsize] at (2.55,0.45) {$\vec{v}_{\text{top-right}}$};
    \node[red, font=\scriptsize] at (2.8,0.1)  {$\vec{v}_{\text{bottom-left}}$};
    \node[red, font=\scriptsize] at (1.8,0.1)  {$\vec{v}_{\text{bottom-right}}$};

    \node[above left] at (C) {$C$};
    \draw[fill=black] (C) circle (0.03);
\end{tikzpicture}

%% file: figure/method/fig_projected_object.tikz.tex
\begin{tikzpicture}[scale=1.8, every node/.style={font=\scriptsize}]
    \coordinate (C) at (0.6,1.0);

    \coordinate (TL) at (5.0,-1.1);
    \coordinate (TR) at (5.2,-2.2);
    \coordinate (BR) at (2.7,-1.8);
    \coordinate (BL) at (3.0,-1.0);
    \coordinate (CBB) at (3.85,-1.4);

    \draw[-, black, line width=0.4pt, dash pattern=on 2pt off 1.5pt] (TL) -- (TR) -- (BR) -- (BL) -- cycle; 
    \draw[-, black, line width=0.4pt, dash pattern=on 2pt off 1.5pt] (C) -- (TL);
    \draw[-, black, line width=0.4pt, dash pattern=on 2pt off 1.5pt] (C) -- (TR);
    \draw[-, black, line width=0.4pt, dash pattern=on 2pt off 1.5pt] (C) -- (BR);
    \draw[-, black, line width=0.4pt, dash pattern=on 2pt off 1.5pt] (C) -- (BL);

    \draw[-, black] (C) -- ($(C)!0.135!(TL)$);
    \draw[-, black] (C) -- ($(C)!0.095!(TR)$);
    \draw[-, black] (C) -- ($(C)!0.18!(BR)$);
    \draw[-, black] (C) -- ($(C)!0.23!(BL)$);
    \draw[-, black] ($(C)!0.135!(TL)$) -- ($(C)!0.095!(TR)$) -- ($(C)!0.18!(BR)$) -- ($(C)!0.23!(BL)$) -- cycle;

    \draw[red] (C) -- (CBB);

    \node[above left] at (C) {$C$};
    \draw[fill=black] (C) circle (0.03);

    \node[below right] at (CBB) {$O$};
    \draw[fill=black] (CBB) circle (0.03);
\end{tikzpicture}

%% file: figure/method/fig_point_plane_distance.tikz.tex
\begin{tikzpicture}[scale=1.8, every node/.style={font=\scriptsize}]

    \coordinate (O) at (0,0);
    \filldraw[black] (O) circle (0.03);

    \coordinate (A) at (-1.5, 1.0); 
    \coordinate (B) at (1.5, 1.0);  
    \coordinate (C) at (1.5, -1.0); 
    \coordinate (D) at (-1.5, -1.0); 

    \draw[thick, red] (A) -- (B) -- (C) -- (D) -- cycle;


    \draw[<->] (0,- 0.1) -- (0, -0.9) node[midway, right] {\scriptsize{$\text{dist}(\vec{O}, \Pi_{\text{bottom}})$}};
    \draw[<->] (0, 0.1) -- (0, 0.9) node[midway, right] {\scriptsize{$\text{dist}(\vec{O}, \Pi_{\text{top}})$}};
    \draw[<->] (-1.3, 0) -- (-0.1, 0) node[midway, above] {\scriptsize{$\text{dist}(\vec{O}, \Pi_{\text{left}})$}};
    \draw[<->] (0.1, 0) -- (1.3, 0) node[midway, above] {\scriptsize{$\text{dist}(\vec{O}, \Pi_{\text{right}})$}};

    \node[red, font=\scriptsize] at (0.0,1.1) {$\Pi_{\text{top}}$};
    \node[red, font=\scriptsize] at (0.0,-1.1) {$\Pi_{\text{bottom}}$};
    \node[red, font=\scriptsize, rotate=90] at (-1.7,0.0) {$\Pi_{\text{left}}$};
    \node[red, font=\scriptsize, rotate=90] at (1.7,0.0) {$\Pi_{\text{right}}$};

\end{tikzpicture}